\PassOptionsToPackage{dvipsnames}{xcolor}
\documentclass{article} 
\usepackage{collas2023_conference,times}
\usepackage{easyReview}
\usepackage{color,soul}

\usepackage{amsmath,amsfonts,bm}

\def\eqref#1{equation~\ref{#1}}

\def\1{\bm{1}}

\DeclareMathAlphabet{\mathsfit}{\encodingdefault}{\sfdefault}{m}{sl}
\SetMathAlphabet{\mathsfit}{bold}{\encodingdefault}{\sfdefault}{bx}{n}

\usepackage{hyperref}
\hypersetup{
    colorlinks=true,
    linkcolor=red,
    filecolor=magenta,
    urlcolor=blue,
    citecolor=purple,
    pdftitle={Overleaf Example},
    pdfpagemode=FullScreen,
    }

\usepackage{times}
\usepackage{epsfig}

\usepackage{graphicx}

\usepackage{tikz}
\usepackage{comment}
\usepackage{amsmath,amssymb} 

\usepackage[subrefformat=parens,labelformat=parens,caption=false]{subfig}
\usepackage{graphicx}
\usepackage{booktabs}
\usepackage{multicol}
\usepackage{multirow}
\usepackage{algorithm2e}
\usepackage{enumitem}
\usepackage[normalem]{ulem}

\newcommand{\ourmodel}{\textit{PlaStIL}\@\xspace}
\newcommand{\ourmodelNospace}{\textit{PlaStIL}}

\newcommand{\deepslda}{$DSLDA$\@\xspace}
\newcommand{\deesil}{$DeeSIL$\@\xspace}
\newcommand{\spbm}{$SPB$-$M$\@\xspace}

\definecolor{armygreen}{rgb}{0.29, 0.33, 0.13}

\title{PlaStIL: Plastic and Stable Exemplar-Free Class-Incremental Learning}

\author{
Grégoire Petit\textsuperscript{1,2}, 
Adrian Popescu\textsuperscript{1}, 
Eden Belouadah\textsuperscript{3}, 
David Picard\textsuperscript{2},
Bertrand Delezoide\textsuperscript{3}\\
 \textsuperscript{1}Université Paris-Saclay, CEA, LIST, F-91120, Palaiseau, France\\
 \textsuperscript{2}LIGM, Ecole des Ponts, Univ Gustave Eiffel, CNRS, Marne-la-Vallée, France\\
 \textsuperscript{3}Datakalab, 114 Bd Malesherbes, 75017 Paris, France\\
 \textsuperscript{4}Amanda, 34 Avenue Des Champs Elysées, F-75008, Paris, France\\
{\tt\small \{gregoire.petit, adrian.popescu\}@cea.fr,eb@datakalab.com}\\
{\tt\small david.picard@enpc.fr,bertrand.delezoide@amanda.com}
}

\collasfinalcopy 

\begin{document}

\maketitle
\begin{abstract}
Plasticity and stability are needed in class-incremental learning in order to learn from new data while preserving past knowledge.
Due to catastrophic forgetting, finding a compromise between these two properties is particularly challenging when no memory buffer is available.
Mainstream methods need to store two deep models since they integrate new classes using fine-tuning with knowledge distillation from the previous incremental state.
We propose a method which has similar number of parameters but distributes them differently in order to find a better balance between plasticity and stability. 
Following an approach already deployed by transfer-based incremental methods, we freeze the feature extractor after the initial state.
Classes in the oldest incremental states are trained with this frozen extractor to ensure stability.
Recent classes are predicted using partially fine-tuned models in order to introduce plasticity. 
Our proposed plasticity layer can be incorporated to any transfer-based method designed for exemplar-free incremental learning, and we apply it to two such methods. 
Evaluation is done with three large-scale datasets. 
Results show that performance gains are obtained in all tested configurations compared to existing methods. 

\end{abstract}


\section{Introduction}
\label{sec:intro}
Class-incremental learning (CIL) enables the adaptation of artificial agents to dynamic environments in which data occur sequentially.
CIL is particularly useful when the training process is performed under memory and/or computational constraints~\citep{masana2021_study}.
However, it is really susceptible to catastrophic forgetting, which refers to the tendency to forget past information when learning new data~\citep{kemker2018measuring,mccloskey:catastrophic}. 
Most recent CIL methods~\citep{douillard2020podnet,hou2019_lucir,javed2018_revisiting,rebuffi2017_icarl,wu2019_bic} use fine-tuning with knowledge distillation~\citep{hinton2015_distillation} from the previous model to preserve past information.
Knowledge distillation has been progressively refined~\citep{hou2019_lucir,javed2018_revisiting,wu2021striking,sdc_2020,zhou2019_m2kd} to improve CIL performance.
An alternative approach to CIL is inspired by transfer learning~\citep{DBLP:conf/cvpr/RazavianASC14}. 
These methods use a feature extractor which is frozen after the initial CIL state~\citep{belouadah2018_deesil,hayes2020_remind,hayes2020_deepslda,rebuffi2017_icarl}. 
They become competitive in exemplar-free CIL, a difficult setting due to a strong effect of catastrophic forgetting~\citep{masana2021_study}.
The main challenge is to find a good plasticity-stability balance because fine-tuning methods favor plasticity, while transfer-based methods only address stability. 

\begin{figure*}[!htbp]
	\centering
	\includegraphics[height=0.28\linewidth]{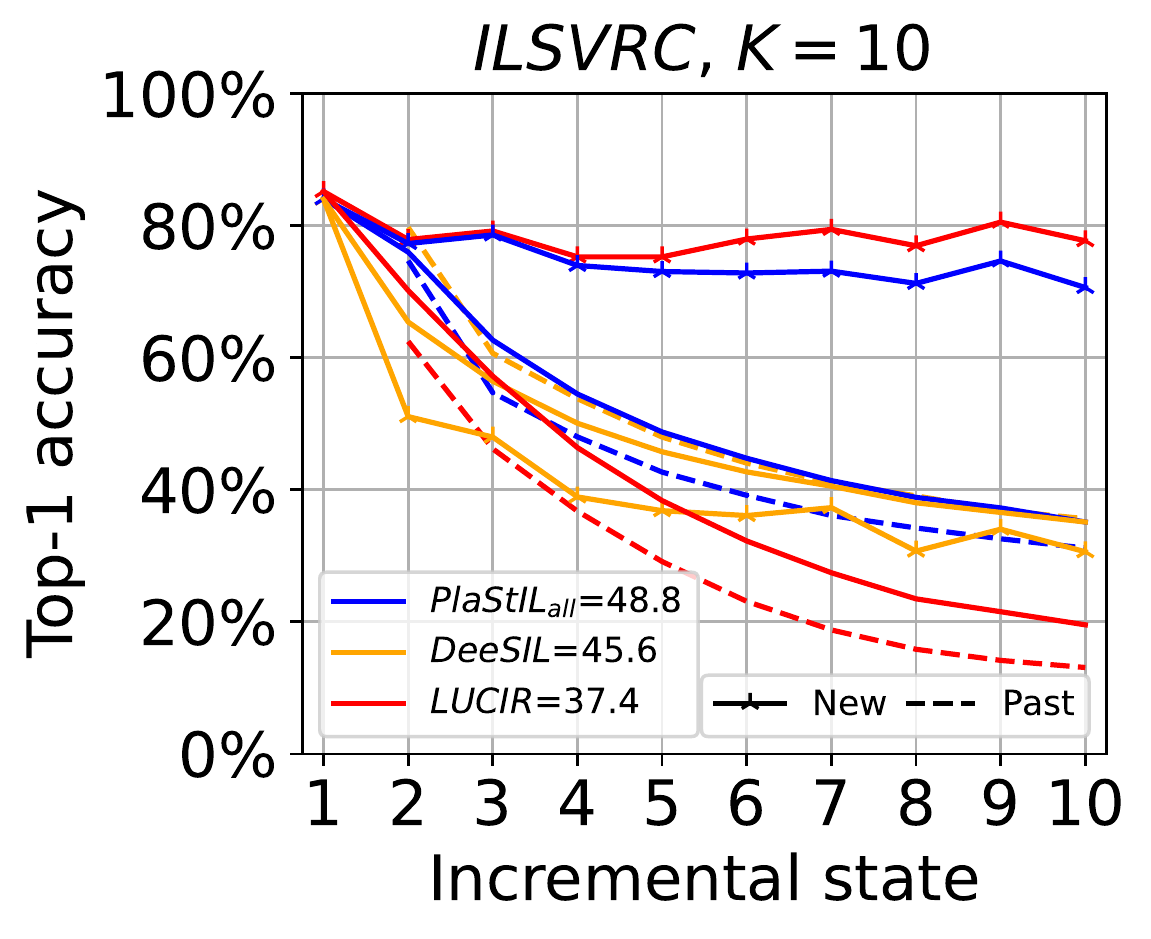}
	\includegraphics[height=0.28\linewidth]{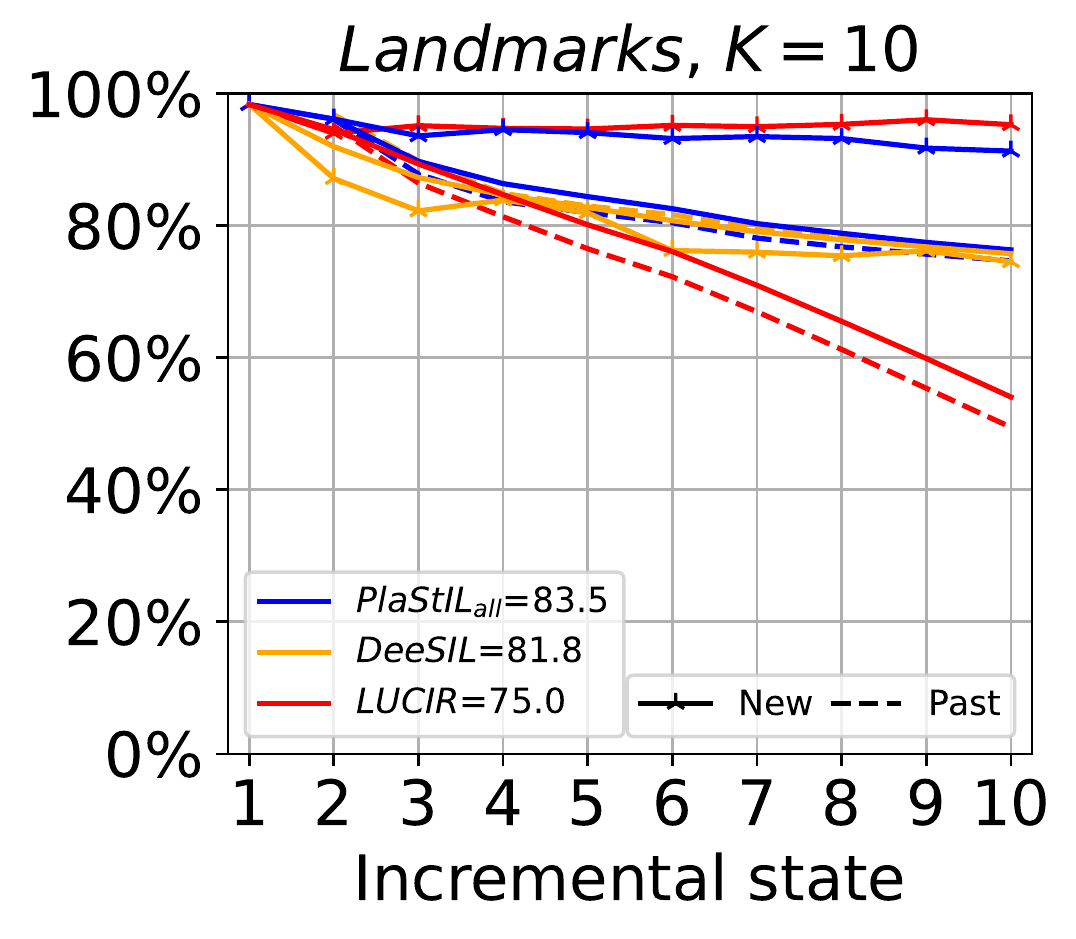}
	\includegraphics[height=0.28\linewidth]{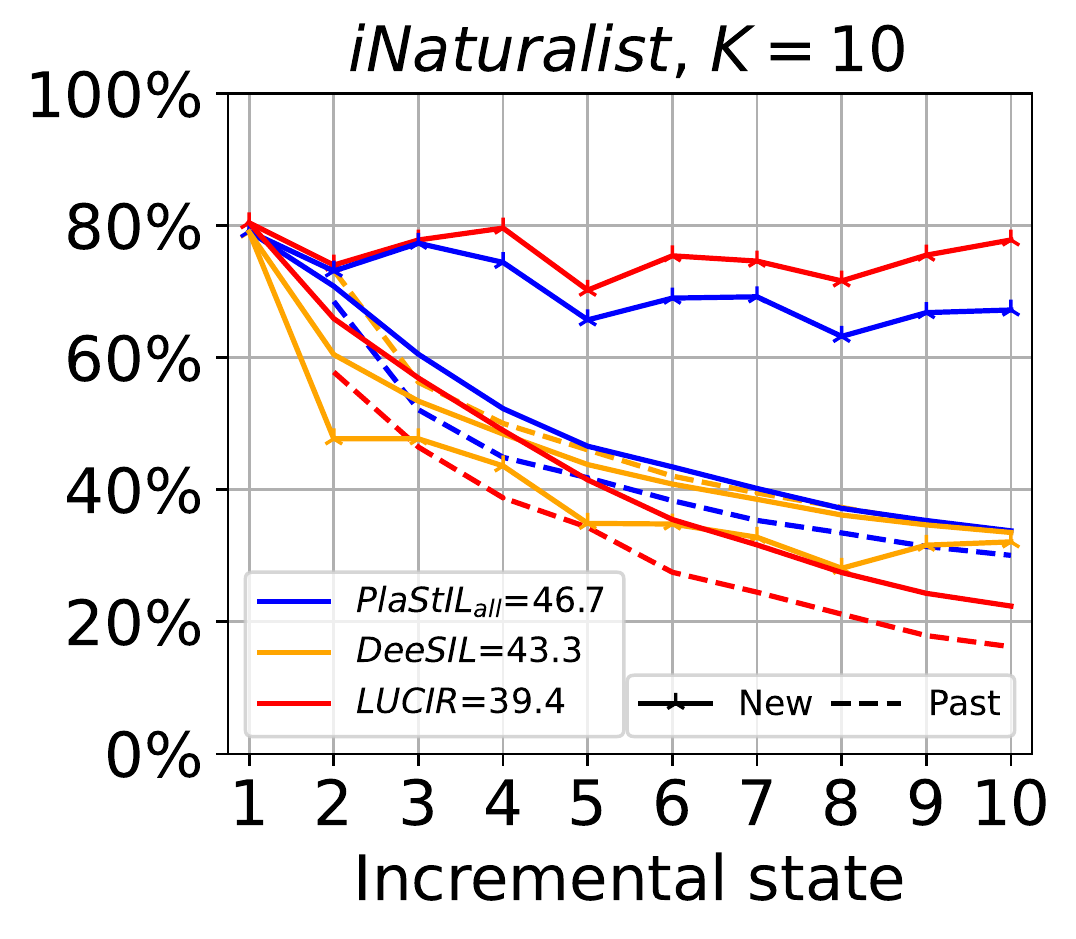}
	
	\caption{Accuracy of past and new classes in exemplar-free CIL for three large-scale datasets with $K=10$  incremental states. $LUCIR$~\citep{hou2019_lucir} uses distillation to preserve past knowledge and favors plasticity. $DeeSIL$~\citep{belouadah2018_deesil} transfers features from the initial frozen model to all subsequent states and focuses on stability. \ourmodel offers a better plasticity-stability balance. Note that the proportion of past classes increases as the incremental process advances and so does their weight in global accuracy.  
	}
	\label{fig:teaser}
\end{figure*}

In this work, we tackle exemplar-free CIL (EFCIL) by combining the two types of approaches described above.
Building on the strong performance of transfer-based methods~\citep{belouadah2021_study,hayes2020_deepslda}, we introduce a plasticity component by partially fine-tuning models for recent classes. 
The results from Figure~\ref{fig:teaser} show that our method gives a better global accuracy compared to $DeeSIL$~\citep{belouadah2018_deesil} and $LUCIR$~\citep{hou2019_lucir}, two representative methods focused on stability and plasticity, respectively.
Accuracy is presented separately for past and new classes for existing methods to examine the plasticity-stability balance offered by each method. 
$LUCIR$ has optimal plasticity (best accuracy of new classes), while $DeeSIL$ has optimal stability (best accuracy for past classes). 
However, the performance of both methods is strongly degraded on the complementary dimensions.
Our method is close to $LUCIR$ in terms of plasticity and to $DeeSIL$ in terms of stability. 
Consequently, it ensures a better balance between these two properties of EFCIL.

\ourmodel is inspired by transfer learning but adds a partial fine-tuning component to boost plasticity.
It is applicable to any transfer-based method and we exemplify it with \deepslda~\citep{hayes2020_deepslda} and $DeeSIL$~\citep{belouadah2018_deesil}.
We introduce a hybrid classification layer which combines classification weights learned with the initial model for past classes and with the fine-tuned models for recent classes.
We evaluate the proposed approach on three datasets which contain 1000 classes each.
The number of incremental states is varied to assess the robustness of the tested methods.
Results show that performance gains are obtained by adding the proposed plasticity component to transfer-based methods.
Equally interesting, important performance improvements are obtained over distillation-based methods, which are the mainstream methods deployed to tackle CIL~\citep{hou2019_lucir,javed2018_revisiting,rebuffi2017_icarl,smith2021always,wu2021striking}.
We will open-source the code to facilitate reproducibility.

\section{Related Work}
\label{sec:related}
Incremental learning is interesting when artificial agents need to learn under memory or computational constraints~\citep{masana2021_study,parisi2019_continual,rebuffi2017_icarl}.
The main challenge in CIL is to tackle the catastrophic forgetting phenomenon \citep{kemker2018measuring,mccloskey:catastrophic}. 
A suitable balance between plasticity and stability of the learned models is sought~\citep{mermillod2013_stability_plasticity}.
Plasticity and stability are needed in order to accommodate new data and preserve previously learned knowledge, respectively~\citep{chaudhry2018_rwalk}.
As noted in a recent survey~\citep{masana2021_study}, a large majority of CIL-related works use a memory buffer which stores samples of past classes in order to improve overall performance.
Replaying these samples facilitates the preservation of past knowledge, thus making the incremental learning process akin to imbalanced learning~\citep{belouadah2021_study}.
However, the assumption that past samples are available is strong and limits the applicability of CIL.
A growing research effort was devoted to exemplar-free CIL (EFCIL)~\citep{belouadah2021_study,masana2021_study}.
The plasticity-stability dilemma is particularly challenging without memory since the effects of catastrophic forgetting are stronger in this case~\citep{belouadah2020_siw,rebuffi2017_icarl,smith2021always,wu2021striking}.

{Survey papers such as}~\citep{lange2019,masana2021_study} {analyze different types of continual learning methods which are usable in exemplar-free CIL. Parameter-isolation methods, such as} $HAT$~\citep{hat_2018} or $PackNet$~\cite{packnet_2018} {were designed for task-incremental learning, a setting in which the task ID is known at inference time. They learn task-specific masks to reduce catastrophic forgetting. However, they are impractical in task-agnostic scenarios since the simultaneous evaluation of all tasks is not possible and specific forward passes are needed for each task} \citep{masana2021_study}.
{Regularization-based methods are a popular solution to EFCIL and they fall into two subcategories, namely data-focused or prior-focused approaches. Existing works}~\citep{lange2019,masana2021_study,petit2023fetril} {showed that data-focused methods outperform prior-focused on in EFCIL scenarios. Consequently, we discuss data-focused methods and use representative examples of them in experiments.}
According to~\citet{belouadah2021_study,masana2021_study}, most methods update learned models in each IL state using fine-tuning for plasticity and different flavors of knowledge distillation~\citep{hinton2015_distillation} for stability.
Alternatively, a few works~\citep{belouadah2018_deesil,dhamija2021self,hayes2020_remind,hayes2020_deepslda} use an initial representation throughout the incremental process.
We discuss the merits and limitations of both approaches below and position our contribution with respect to them.

Distillation-based methods are inspired by $LwF$~\citep{li2016_lwf}, an adaptation of knowledge distillation \citep{hinton2015_distillation} to an incremental context. 
The authors of~\cite{dhar2018_lwm} add an attention mechanism to the distillation loss to preserve information from past classes and obtain an improvement over $LwF$.
Since its initial use for exemplar-based CIL in $iCaRL$~\citep{rebuffi2017_icarl}, distillation was refined and complemented with other components to improve the plasticity-stability compromise. 
$LUCIR$~\citep{hou2019_lucir} applies distillation on feature vectors instead of raw classification scores to preserve the geometry of past classes, and an inter-class separation to maximize the distances between past and new classes. $LwM$~\citep{dhar2018_lwm} adds an attention mechanism to the distillation loss to preserve information from base classes. 
An interesting solution is proposed in~\citet{sdc_2020}, where the feature drift between incremental steps is estimated based on features of samples associated to new classes.
However, this method has a large footprint since it needs a large multi-layer perceptron and also stores past features to learn the transformation.
A feature transformation method is designed for task-incremental learning and adapted for CIL by predicting the task associated to each test sample~\citep{kumar2021_efficient}. 
The authors of~\citet{smith2021always} combined feature drift minimization and class separability to improve distillation.
The authors of~\citet{wu2021striking} proposed an approach which stabilizes the fine-tuned model, adds reciprocal adaptive weights to weigh past and new classes in the loss, and introduces multi-perspective training set augmentation.
They reported important gains in exemplar-free CIL compared to $LUCIR$~\citep{hou2019_lucir} and $SDC$~\citep{sdc_2020} using a protocol in which half of the dataset is available initially~\citep{hou2019_lucir}. 
Distillation is widely used but a series of studies question its usefulness~\citep{belouadah2021_study,masana2021_study,prabhu2020gdumb}, especially for large-scale datasets.
One explanation for the lack of scalability of distillation is that inter-class confusion becomes too strong when the number of past classes is high.
Another challenge is that distillation needs to store the previous deep model to preserve past knowledge. 
The total footprint of these methods is the double of the footprint of the backbone model used.

A second group of methods learns a deep representation in the initial state and uses it as feature extractor throughout the CIL process.
They are inspired by transfer learning~\citep{tan2018survey} and favor stability since the initial representation is frozen.
Usually, these methods learn shallow external classifiers on top of the initial deep representation.
The nearest class mean ($NCM$)~\citep{mensink2013distance} was introduced in \citet{rebuffi2017_icarl}, linear SVCs~\citep{cortes1995support} were used in~\citet{belouadah2018_deesil,petit2023fetril} and extreme value machines~\citep{rudd2017extreme} were tested by~\citet{dhamija2021self}. 
$REMIND$~\citep{hayes2020_remind} {uses a vector quantization technique to save compressed image representations which are later reconstructed for memory consolidation by training model tops. The main difference with our proposal is that the past is represented via compressed image representations instead of model tops.}
$DSLDA$ \citep{hayes2020_deepslda} updates continuously a class-specific mean vector and a shared covariance matrix.
The predicted label is the one having the closest Gaussian in the feature space defined by these vectors and matrix. 
These methods are simple and suited for exemplar-free CIL, particularly for large-scale datasets where they outperform distillation-based methods~\citep{belouadah2021_study,masana2021_study}.
Equally important, they only use the initial model and thus have a smaller footprint.
Their main drawbacks are the genericity of the initial representation and the sensitivity to strong domain variations. 
Their performance drops if a small number of classes is initially available~\citep{belouadah2021_study} and if the incremental classes have a large domain shift with classes learned initially~\citep{lange2019}.
{The robustness of the initial representation can be improved is an assumption is made that a model pretrained with a large amount of data is available and that its features are transferable to the incremental datasets}~\citep{hayes2020_remind}. 
$ESN$~\citep{esn_2023} {is an interesting method which was proposed very recently, and makes this assumption, similarly to} $REMIND$ and \deepslda. $ESN$ { leverages a pretrained transformer model, trains classifiers per state and then merges them by combining a temperature-controlled energy metric, an anchor-based energy self-normalization strategy and a voting-based inference augmentation strategy to ensure impartial and robust EFCIL predictions. Here, we experiment without a large pretrained model in order to cover cases where there is a large domain drift between the pretraining and the incremental datasets.}


\section{Proposed Method}
\label{sec:method}
\subsection{Problem Formalization}
The CIL process is divided in $K$ states, with $n$ classes learned in each state.
In EFCIL, no past data can be stored for future use.
The predictions associated to observed classes are noted $p$.
We write the structure of a deep model as:
\begin{equation}
\mathcal{M} = \{\mathcal{B} , \mathcal{T} , \mathcal{W}    
\label{eq:model} \} 
\end{equation} 
with:
$\mathcal{M}$ - the full model;
$\mathcal{B}$ - the model base which includes the initial layers;
$\mathcal{T}$ - the model top which includes the subsequent layers up to the classification one;
$\mathcal{W}$ - the classification layer which provides class predictions.

Assuming that the CIL process includes $K$ states, the objective is to learn $K$ models in order to incorporate all classes which arrive sequentially.
The incremental learning process can be written as: 
\begin{equation}
\mathcal{M}_1 \rightarrow \mathcal{M}_2 \rightarrow ... \rightarrow \mathcal{M}_k \rightarrow ... \rightarrow  \mathcal{M}_{K-1} \rightarrow \mathcal{M}_K    
\end{equation}

Each incremental model needs to integrate newly arrived data, while also preserving past knowledge.
Assuming that the current state is $k \geq 2$, the majority of existing CIL methods~\citep{castro2018_e2eil,hou2019_lucir,rebuffi2017_icarl,smith2021always,wu2021striking,wu2019_bic} fine-tunes the entire current model $\mathcal{M}_k$ by distilling knowledge from $\mathcal{M}_{k-1}$.
We note $w_k^1$ to $w_k^{(k-1) \times n}$ the classifier weights of the past states $1$ to $k-1$, and $w_k^{(k-1) \times n+1}$ to $w_k^{k \times n}$ the classifier weights of the new state $k$.

Their classification layer is written as:
\begin{equation}
\hspace{-0.5em}
\mathcal{W}_k^{ft} =\{w_k^1, ..., w_k^n, ..., w_k^{(k-1) \times n}, w_k^{(k-1) \times n+1}, ..., w_k^{k \times n} \}  
\label{eq:updated}
\end{equation}
They are all trained using $\mathcal{T}_k$, the model top learned in the $k^{th}$ state.
The $\mathcal{W}_k^{ft}$ layer is biased toward new classes since it is learned with all samples from the current state, but only with the representation of past classes stored in $\mathcal{M}_{k-1}$~\citep{hou2019_lucir,rebuffi2017_icarl,wu2019_bic}.
This group of methods focuses on CIL plasticity at the expense of stability~\citep{belouadah2021_study,masana2021_study}.

Transfer-based methods~\citep{belouadah2018_deesil,hayes2020_remind,hayes2020_deepslda,petit2023fetril} freeze the feature extractor $\mathcal{F}_1 = \{\mathcal{B}_1 , \mathcal{T}_1$\} after the initial non-incremental state.
All the classes observed during the CIL process are learned with $\mathcal{F}_1$ as features extractor.
The classification layer can be written as:

\begin{equation}
\hspace{-0.5em}
\mathcal{W}_1^{fix} =\{w_1^1, ..., w_1^n, ..., w_1^{(k-1) \times n}, w_1^{(k-1) \times n+1}, ..., w_1^{k \times n} \} 
\label{eq:init}
\end{equation}

All classifier weights from Eq.~\ref{eq:init} are learned with image features provided by $\mathcal{F}_1$, the feature extractor learned initially, inducing a bias toward initial classes. 
It is suboptimal for classes learned in states $k \geq 2$ because their samples were not used to train $\mathcal{F}_1$. 
These methods focus on CIL stability  at the expense of  plasticity~\citep{masana2021_study}.

\subsection{\ourmodel Description}
\label{subsec:method}
\ourmodel is motivated by recent studies which question the role of distillation in CIL, particularly for large-scale datasets~\citep{belouadah2021_study,masana2021_study,prabhu2020gdumb}. 
Instead of $\mathcal{M}_{k-1}$ needed for distillation, \ourmodel uses two or more model tops $\mathcal{T}$ which have an equivalent number of parameters at most.
\ourmodel is inspired by feature transferability works~\citep{neyshabur2020being,yosinski2014transferable} which show that higher layers of a model, included in $\mathcal{T}$, are the most important for successful transfer learning.
Consequently, the initial layers 
$\mathcal{B}$ are frozen and shared throughout the incremental process.
A combination of model tops which includes $\mathcal{T}_1$, the one learned in the first incremental state, and those of the most recent state(s) is used in \ourmodel.
Similar to transfer-based CIL methods~\citep{belouadah2018_deesil,hayes2020_remind,hayes2020_deepslda,petit2023fetril}, $\mathcal{T}_1$ ensures stability for classes first encountered in past states for which a dedicated top model is not available. 
Different from existing methods, model top(s) are available for the most recent state(s), thus improving the overall plasticity of \ourmodel. 
The number of different model tops which can be stored instead of $\mathcal{M}_{k-1}$ depends on the number of higher layers which are fine-tuned in each incremental state.
The larger the number of layers in $\mathcal{T}$, the larger its parametric footprint is and the lower the number of storable model tops will be.

\begin{figure*}[t]
	\centering
    \includegraphics[width=0.98\linewidth]{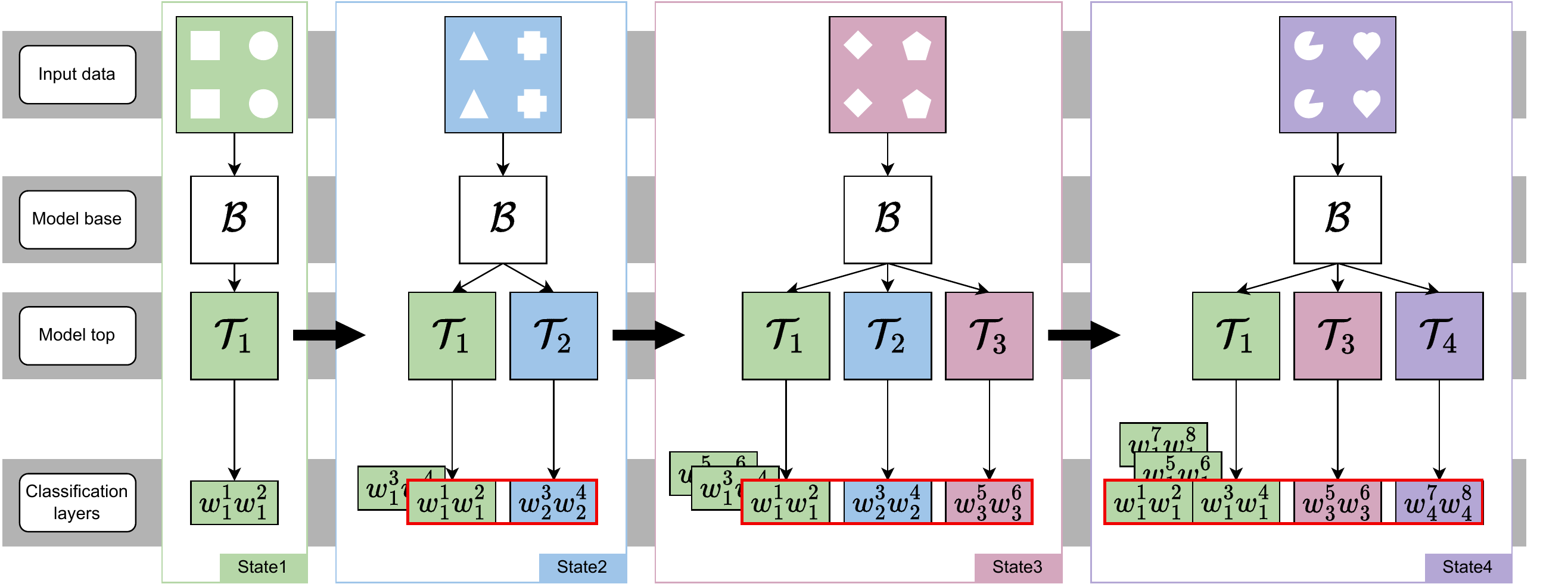}
	\caption{\ourmodel overview using a toy example with $K=4$ CIL states and $n=2$ new classes learned per state. The global memory footprint is equivalent to that of distillation-based methods, but this memory is used differently. We assume that a model base and at most three model tops can be used. A base $\mathcal{B}$, is learned in the initially and then frozen, as is $\mathcal{T}_1$ which is needed to ensure stability. Initial classifier weights are trained using $\mathcal{T}_1$ in each state and reserved for future use. Classifier weights which are actually used in each state are highlighted in red. In state 4, the recent model tops ($\mathcal{T}_4$ and $\mathcal{T}_3$) are included to ensure plasticity. Classifier weights $w_4^7$ and $w_4^8$, associated to the new classes from state 4 are learned with features provided by $\mathcal{T}_4$. $w_3^5$ and $w_3^6$ were learned with $\mathcal{T}_3$ features in state 3, when they were new. $w_1^1$ to $w_1^4$ were learned with $\mathcal{T}_1$ features, in states 1 and 3. $w_1^5$ to $w_1^8$ are reserved for future use. $\mathcal{T}_2$ is discarded to keep the total memory footprint of \ourmodel bounded. \textit{Best viewed in color.}
	}
	\label{fig:overview}
\end{figure*}

The method is illustrated in Figure~\ref{fig:overview} with a toy example which includes $K=4$ IL states, with $n=2$ new classes per state and which assumes that up to three model tops can be stored. 
Up to the third state, \ourmodel stores a model top per state and corresponding classifier weights are learned for each model top.
In the fourth state, one of the model tops needs to be removed in order to keep the parameters footprint bounded. 
Consequently, $\mathcal{T}_2$ is removed and the initial model top $\mathcal{T}_1$ is used. 
Note that $\mathcal{T}_1$ is used to learn classifier weights for all classes when they occur initially.
These initial weights are stored for usage in later incremental states in order to cover all past classes for which dedicated model tops cannot be stored. 
The storage of the initial weights generates a small parameters overhead but its size is small and does not increase over time.
If the classifier weights are $d$-dimensional, the number of supplementary parameters is $n \times d$.
Moreover, this overhead can be easily compensated by the choice of number of parameters in $\mathcal{T}$ and the number of such model tops which are stored.
In Figure~\ref{fig:overview}, initial classifier weights $w_1^3$ and $w_1^4$ are first learned in state 2 but only used in state 4, when $\mathcal{T}_2$ is no longer available. 
In state 4, $\mathcal{T}_1$ is used for classes which first occurred in states 1 and 2 (classifiers weights $w_1^1$ to $w_1^4$).
$\mathcal{T}_3$ is reused along with its classifier weights $w_3^5$ and $w_3^6$, learned for the classes which were learned in state 3. 
Finally, classifier weights of new classes $w_4^7$ and $w_4^8$ are learned with $\mathcal{T}_4$.
This results into a hybrid classification weights layer which is defined as:

\begin{equation}
\hspace{-0.5em}
\mathcal{W}_k^{hyb} =\{
\textcolor{ForestGreen}{w_1^1, ..., w_1^{j \times n}},
...,
\textcolor{RoyalBlue}{w_{j+1}^{j \times n+1}, ..., w_{j+1}^{(j+1) \times n}},
...
,
\textcolor{Purple}{w_k^{(k-1) \times n+1}, ..., w_k^{k \times n}} \}
\label{eq:hybrid}
\end{equation}

where we assume that $k-j+1$ models can be stored, with $2 \leq j \leq k$; the blocks of classes learned with features from different model tops are color coded.

In Equation~\ref{eq:hybrid}, classifier weights of the first $j$ incremental states are learned with the features provided by initial model top $\mathcal{T}_1$.
Those of  the most recent states ($j+1$ to $k$) are learned with features provided by model tops $\mathcal{T}_{j+1}$ to $\mathcal{T}_k$. 
An advantage of the layer from Equation~\ref{eq:hybrid} is that it ensures a good balance between stability, via $\mathcal{T}_1$ and plasticity, via $\mathcal{T}_{j+1}$ to $\mathcal{T}_k$.
The number of storable model tops varies inversely with the number of layers that they include. 
We report results with three top depths in Section~\ref{sec:evaluation}.
A choice between internal and external classifiers has to be made for the implementation of this classification layer. 
Experiments from~\citet{belouadah2021_study} indicate that external classifiers are easier to optimize when transferring features from $\mathcal{M}_1$ to subsequent incremental states. 

\ourmodel {is primarily intended for a CIL scenario under the assumption that the parametric budget should not increase over time. If memory is allowed to grow over time, the method could store a larger number of model tops. The model top creation and removal policy could be adapted depending on the continual learning scenario being explored. For instance, if classes were grouped semantically, as it is the case in task-incremental learning, it might be better to create model tops for states which include classes which are most dissimilar from those of the initial model. Another interesting scenario assumes that past classes can be revisited, with new samples of them arriving later in the incremental process. In this case, model tops could be created for the current state if the amount of samples for revisited classes is smaller than those associated to existing tops. Such a top creation policy would be based on the assumption that the less revisiting there is, the more likely a new top would be due to the fact that the current state includes more novelty. In practice, the total number of stored model tops will depend on the total budget available on the device. When the memory budget is reached, one of the selection strategies listed above can be applied depending on the characteristics of the continual learning process. In Subsection}~\ref{subsec:analysis}, {we show that the creation of model tops for the most recent states is a good solution for class incremental learning. While interesting, the adaptation of} \ourmodel {to other continual learning scenarios is out of the immediate scope and is thus left for future work.} 


\section{Experiments}
\label{sec:evaluation}
We evaluate \ourmodel with three large-scale datasets designed for different visual tasks. 
We compare it to a representative set of EFCIL methods. 
We vary the total number of states $K$ using $K\in\{5, 10, 20\}$ because the length of the CIL process has strong effects on EFCIL performance~\citep{masana2021_study}.
The evaluation metric is the top-1 accuracy averaged over all incremental states.
Following a common practice in CIL~\citep{castro2018_e2eil,hou2019_lucir,wu2019_bic}, the performance on the initial state is excluded because it is not incremental.

\subsection{Datasets}
\label{subsec:datasets}
We thus select large-scale datasets which provide a more realistic scenario for evaluation compared to medium-scale ones which are still used~\citep{smith2021always,wu2021striking}.
We run experiments with:
\begin{itemize}[noitemsep,leftmargin=*]
	\item $ILSVRC$~\citep{olga2015_ilsvrc}  - the well-known subset of ImageNet~\citep{deng2009_imagenet} built for the eponymous competition and also used in CIL~\citep{castro2018_e2eil,hou2019_lucir,rebuffi2017_icarl,wu2019_bic}. The training and testing sets are composed of 1,231,167 and 50,000 images, respectively.
	\item $Landmarks$ - a subset of a landmarks recognition dataset~\citep{noh2017_landmarks} which includes a total of over 30000 classes. We select the 1000 classes having the largest number of images. The training and testing sets are composed of 374,367 and 20,000 images, respectively.
	\item $iNaturalist$ - a subset of the dataset used for the iNaturalist challenge~\citep{van2018inaturalist}. The full version includes 10000 fine-grained classes for natural species. We sample 1000 classes from different super-categories to obtain a diversified subset. The training and testing sets are composed of 300,000 and 10,000 images, respectively.
\end{itemize}


\subsection{State-of-the-art methods}
\label{subsec:baselines}
We compare \ourmodel with the following existing methods:
\begin{itemize}[noitemsep,leftmargin=*]
    \item $LwF$~\citep{rebuffi2017_icarl} - is a CIL version of the initial method from~\citet{li2016_lwf}. It tackles forgetting using a distillation loss.
    \item $SIW$~\citep{belouadah2020_siw}  - uses a vanilla \textit{FT} backbone and tackles catastrophic forgetting by reusing the past classifiers learned when these classes were first learned.
    \item $LUCIR$~\citep{hou2019_lucir}  - adapts distillation to feature vectors instead of raw scores to preserve the geometry of past classes and also pushes for inter-class separation. Note that while initially proposed for CIL with memory, this method showed strong performance in EFCIL too~\citep{belouadah2021_study}. 
    \item \spbm~\citep{wu2021striking} - is a recent method which focuses on balancing plasticity and stability in exemplar-free CIL. We report results with the multi-perspective variant, which has the best overall performance in~\citet{wu2021striking}. \spbm provides very competitive performance compared to $LUCIR$ when half of the dataset is initially available. 
    \item $PASS$~\citep{zhu2021_pass} - {uses prototypes of past classes in combination with distillation in order to counter catastrophic forgetting.} 
     \item $REMIND$~\cite{hayes2020_remind} - {encodes knowledge about past classes by storing compressed image representations. It is compared with our method by allocating the amount of storage used for model tops in our method to compressed representations of past samples.}   
    \item $FeTrIL$~\citep{petit2023fetril} - is a very recent method which focuses on generating pseudo-features for past classes by using a geometric translation of features of similar new classes. 
    \item \deepslda~\citep{hayes2020_deepslda}  - is based on Gaussian functions defined in the features space by specific mean class vectors and a covariance matrix which is shared among all classes. This method is interesting since its classification layer provides an efficient inter-class separability mechanism.
    \item $DeeSIL$~\citep{belouadah2018_deesil}  - freezes the initial model uses linear SVCs~\citep{cortes1995support} for the final layer, which is trained independently for each state. 
\end{itemize}

The first four methods fine-tune models incrementally.
$REMIND$ trains model tops using a compressed replay buffer and is very relevant for comparison here. 
$DSLDA$ and $DeeSIL$ are transfer-based, and \ourmodel can be applied to them.
\textcolor{black}{They were implemented using their original optimal parameters.
Whenever the original experimental settings were different from the ones used here, the correct functioning of the baselines was carefully checked.
The obtained accuracy was coherent with the results reported in the original papers and/or in comparative studies such as~\citet{belouadah2021_study,masana2021_study}  in all cases.
See details about the reproduced results in the appendix.
}

We experiment with three versions of \ourmodel designed to ensure that its parameters footprint is equivalent to (or lower than) that of distillation-based methods. 
We assume that the incremental process is in the $k^{th}$ state and test:
\begin{itemize}[noitemsep,leftmargin=*]
    \item \ourmodelNospace$_1$ - fine-tunes model tops $\mathcal{T}$ limited to the last convolutional layer of ResNet-18, which includes approximately 21.45\% of the model parameters. Consequently, we can fit $\mathcal{T}_1$, $\mathcal{T}_{k-3}$, $\mathcal{T}_{k-2}$, $\mathcal{T}_{k-1}$ and $\mathcal{T}_{k}$ in memory.
    \item \ourmodelNospace$_2$ - fine-tunes $\mathcal{T}$ which includes the last two convolutional layers of ResNet-18, which includes approximately 42.9\% of the model parameters.  We can fit $\mathcal{T}_1$, $\mathcal{T}_{k-1}$ and $\mathcal{T}_{k}$ in the allowed parameters memory.
    \item \ourmodelNospace$_{all}$ - trains all the layers of the current model in $k^{th}$ state and we can only use $\mathcal{T}_1$ and $\mathcal{T}_k$ in each IL state. 
\end{itemize}

\ourmodel variant test different variants of the compromise between the number of model tops and their depth.
\ourmodelNospace$_1$ fine-tunes only the last convolutional layer of model tops and maximizes the number of such storable models.
\ourmodelNospace$_{all}$  provides optimal transfer since all layers are trained with new data but can accommodate only the current model.
\ourmodelNospace$_2$ provides a compromise between top depth and the number of storable models.

We also provide results for:  (1)  vanilla fine-tuning ($FT$) - a baseline that does not counter catastrophic forgetting at all, and (2) $Joint$ - an upper bound  which consists of a standard training in which all data are available at once. 

\subsection{Implementation}
\label{subsec:implementation}
A ResNet-18 \citep{he2016_resnet} architecture was used as a backbone in all experiments. 
All methods were run with the published optimal parameters and minor adaptation of the codes to unify data loaders: $FT$ \citep{belouadah2020_siw}, $LwF$ \citep{rebuffi2017_icarl}, $LUCIR$ \citep{hou2019_lucir}, $SIW$ \citep{smith2021always}, $DSLDA$ \citep{hayes2020_deepslda}, $DeeSIL$ \citep{belouadah2018_deesil} and $FeTrIL$ \citep{petit2023fetril}. 
\spbm\citep{wu2021striking} has no public implementation and we reimplemented the method. 
We verified its correctness by comparing the accuracy obtained with our implementation (60.1) to the original one (59.7) for $ILSVRC$ split tested by the authors~\citep{wu2021striking}.
All methods were implemented in PyTorch \citep{paszke2019_pytorch}, except for $LwF$ which uses the Tensorflow \citep{abadi2015_tensorflow} implementation of $LwF$ from \citet{rebuffi2017_icarl} because it provides better performance compared to later implementations~\citet{javed2018_revisiting,wu2019_bic}.
The training procedure from \citet{hayes2020_remind} was used to obtain initial models for all transfer-based methods.
These initial models were trained for 90 epochs, with a learning rate of 0.1, a batch size of 128, and a weight decay of $10^{-4}$. 
We used stochastic gradient descent for optimization and divided the learning rate by 10 every 30 epochs.
Detailed parameters are presented in the appendix.

\subsection{Main results}
\label{subsec:results}


The results presented in Table~\ref{tab:main_results} show that all \ourmodel variants improve over the transfer-based methods to which they are added for all tested datasets and CIL configurations.
The gains are generally higher for $K=5$ states, but remain consistent for the $K = \{10, 20\}$.
For instance, \ourmodelNospace$_1$ gains 6.2, 6.4 and 4.4 points for $ILSVRC$ split into 5, 10, 20 states, respectively. 
The best overall performance is obtained with \ourmodelNospace$_1$, followed by \ourmodelNospace$_2$ and \ourmodelNospace$_{all}$ applied on top of $DeeSIL$.
A combination of recent model tops which fine-tune only the last convolutional layer is best here.
{Our method appled to } \deesil {provides best performance for 5 and 10 incremental states. 
Gains are equally interesting for} \deepslda, {and particularly for $K=20$. 
This baseline has better performance than} \deesil {for all three datasets when $K=20$ . The application of} \ourmodel {on top of} \deepslda {leads to slightly better performance compared to the version built on top of} \deesil {for $ILSVRC$ (42.5 vs. 41.9) and $iNaturalist$ (37.4 vs 36.4)}. 
The better behavior of \deepslda~ for longer incremental sequences is explainable since this method features a global inter-class separability component.
In contrast, $DeeSIL$ only separates classes within each state and its discriminative power is reduced when each state includes a low number of classes.

\begin{table}
	\begin{center}
		\resizebox{0.98\textwidth}{!}{
			\begin{tabular}{@{\kern0.5em}lccccccccccccc@{\kern0.5em}}
				\toprule
				\multirow{2}{*}{\textbf{CIL Method}}
				& \multirow{2}{*}{\textbf{{mem on disk}}}
				& \multicolumn{3}{c}{\textit{ILSVRC}}
				&& \multicolumn{3}{c}{\textit{Landmarks}}
				&& \multicolumn{3}{c}{\textit{iNaturalist}}
				\\ \cmidrule(lr){3-5} \cmidrule(lr){7-9} \cmidrule(l){11-13}  
				&
				& \multicolumn{1}{c}{$K$=5}
				& \multicolumn{1}{c}{$K$=10}
				& \multicolumn{1}{c}{$K$=20}
				&
				& \multicolumn{1}{c}{$K$=5}
				& \multicolumn{1}{c}{$K$=10}
				& \multicolumn{1}{c}{$K$=20}
				&
				& \multicolumn{1}{c}{$K$=5}
				& \multicolumn{1}{c}{$K$=10}
				& \multicolumn{1}{c}{$K$=20}
				
				\\ \midrule
				
				$FT$ (\textit{lower bound})                      & {44.59MB} & 26.6 & 18.3 & 12.2 && 31.3 & 21.0 & 13.4 && 25.6 & 17.5 & 11.4 \\  \hline
				$LwF$~\citep{rebuffi2017_icarl} \tiny (CVPR'17)    & {44.59MB} & 24.0 & 21.1 & 17.4 && 36.9 & 34.7 & 28.0 && 23.9 & 21.5 & 16.3 \\
				$SIW$~\citep{belouadah2020_siw} \tiny(BMVC'20)     & {44.59MB} & 38.3 & 35.2 & 26.8 && 66.4 & 55.7 & 41.4 && 38.6 & 30.9 & 17.2 \\ 
				$LUCIR$~\citep{hou2019_lucir}  \tiny(CVPR'19)      & {89.18MB}& 50.4 & 37.4 & 24.4 && 89.5 & 75.0 & 50.5 && 54.9 & 39.4 & 24.8 \\
				$SPB$-$M$~\citep{wu2021striking} \tiny(ICCV'21) & {89.18MB}& 38.9 & 37.3 & 30.4 && 81.6 & 70.4 & 57.1 && 46.7 & 39.6 & 29.8 \\
                $PASS$~\citep{zhu2021_pass} \tiny(CVPR'21)  & {89.18MB} & {39.4} & {35.9} & {29.8} && {65.0} & {55.1} & {42.3} && {48.0} & {40.9} & {31.8} \\		
                $REMIND$~\citep{hayes2020_remind} \tiny(ECCV'20) & {89.18MB}& {52.2} & {44.8} & {35.9} && {83.3} & {77.5} & {72.2} && {50.6} & {39.4} & {31.3} \\
                
                $FeTrIL$~\citep{petit2023fetril} \tiny(WACV'23)& {44.59MB} & 54.7 & 49.1 & \textbf{42.8} && 86.1 & 81.2 & \underline{77.7} && 52.9 & 45.3 & \textbf{38.7} \\ 
			\hline
				$DSLDA$~\citep{hayes2020_deepslda} \tiny(CVPRW'20) & {45.59MB}   & 51.3 & 45.4 & 39.2 && 82.7 & 78.5 & 74.5 && 49.7 & 42.1 & 34.8 \\
				w/ \ourmodelNospace$_{1}$                   & {93.44MB} & 56.8 & 50.1 & 42.2 && 86.8 & 82.1 & 76.1 && 53.8 & 45.6 & 36.6 \\
				w/ \ourmodelNospace$_2$                 & {87.52MB} & \underline{58.3} & \underline{50.6} & \underline{42.5} && 87.8 & 82.1 & 76.0 && 56.1 & 46.2 & 36.9 \\
				w/ \ourmodelNospace$_{all}$           & {91.18MB} & 57.7 & 49.8 & 41.9 && 86.9 & 81.3 & 75.5 && 56.2 & 46.3 & \underline{37.4} \\
			\hline
   
				$DeeSIL$~\citep{belouadah2018_deesil}& {44.59MB}& 52.4 & 45.4 & 37.5 && 87.4 & 80.8 & 73.8 && 52.7 & 43.5 & 33.9 \\ 
				w/ \ourmodelNospace$_1$            & {88.44MB} & 58.6 & \textbf{51.8} & 41.9 && \underline{92.1} & \textbf{86.4} & \textbf{78.1} && 56.8 & \textbf{47.5} & 36.4 \\ 
                w/ \ourmodelNospace$_2$       & {84.52MB} & \textbf{59.2} & 50.2 & 39.7 && \textbf{92.2} & \underline{85.1} & 76.7 && \underline{57.3} & 46.9 & 36.0 \\ 
				w/ \ourmodelNospace$_{all}$      & {89.18MB} & 57.7 & 48.6 & 39.0 && 90.7 & 83.4 & 75.3 && \textbf{58.2} & \underline{47.1} & 35.6 \\ 

			\hline
				\multicolumn{1}{l}{$Joint$ (\textit{upper bound})} &&
				\multicolumn{3}{c}{73.0} &&
				\multicolumn{3}{c}{97.4} &&
				\multicolumn{3}{c}{75.6} \\
				\toprule
			\end{tabular}
		}
	\end{center}
	\caption{Average top-1 accuracy with three numbers of states $K$ per dataset. \ourmodel is applied on top of $DeeSIL$ and \deepslda.  \textbf{Best results - in bold}, \underline{second best - underlined}.}
	\label{tab:main_results}
\end{table}

Distillation-based methods have lower performance compared to transfer-based methods for the tested large-scale datasets.
This result is coherent with previous findings regarding scalability problems of distillation~\citep{belouadah2021_study,masana2021_study,prabhu2020gdumb}.
The difference between \ourmodel applied to $DeeSIL$ and \deepslda and distillation-based methods is very consequent. 
It is in the double-digits range compared to $LUCIR$, the best distillation-based method, for five configurations out of nine tested.
This difference reaches a maximum value of 27.6 top-1 accuracy points for $Landmarks$ with $K=20$ states and a minimum of 2.7 points for the same dataset with $K=5$ states.
\spbm is a recent method which compares very favorably with $LUCIR$ when half of the dataset is allowed in the initial state~\citep{wu2021striking}. 
However, its behavior is globally similar to that of $LUCIR$, with performance gains for $K=20$ states and losses for $K=5$.
This happens because \spbm is more dependent on the representativeness of the initial model compared to $LUCIR$ since it features a strong stability component.
$LwF$ and $SIW$, the two other methods which update models in each incremental state have even lower performance than $LUCIR$ and \spbm.
{The comparison with $REMIND$ is also favorable in all tested configurations. This indicates that, given an identical memory budget, the use of model tops is more effective than the use of a replay buffer made of compressed samples of the past .}
\ourmodel is also better than $FeTrIL$ on 7 configurations out of 9, and ranks similarly (0.3\% and 0.7\%) on the remaining two. It can be explained by the fact that on the long run the pseudo-feature generator of $FeTrIL$ favors the stability. However, on shorter runs, for $K=5$ states, \ourmodel beats it by 4.5\%, 6.1\% and 5.3\%.
All EFCIL methods need a supplementary budget to ensure the plasticity-stability balance.
The takeaway from the comparison of \ourmodel with mainstream methods is that this budget is much better spent on partially fine-tuned model tops than on storing the previous model needed for distillation.

The results per dataset show that $ILSVRC$ and $iNaturalist$ are harder to solve compared to $Landmarks$.
The gains obtained by \ourmodel are smaller for $Landmarks$ since the progress margin is more reduced. 
The performance gap between the evaluated methods and $Joint$ is large.
The fact that the gap widens with the number of incremental states has specific explanations for the two types of methods.
Past knowledge becomes harder to preserve when the number of fine-tuning rounds increases in $LUCIR$, \spbm, $SIW$, $LwF$.
For transfer-based methods, past knowledge is encoded using a weaker feature extractor. 
\ourmodel reduces the gap with $Joint$, but the results reported here show that exemplar-free class-incremental learning  remains a very challenging problem.

\begin{figure*}
\centering
    {{\includegraphics[height=.3\linewidth]{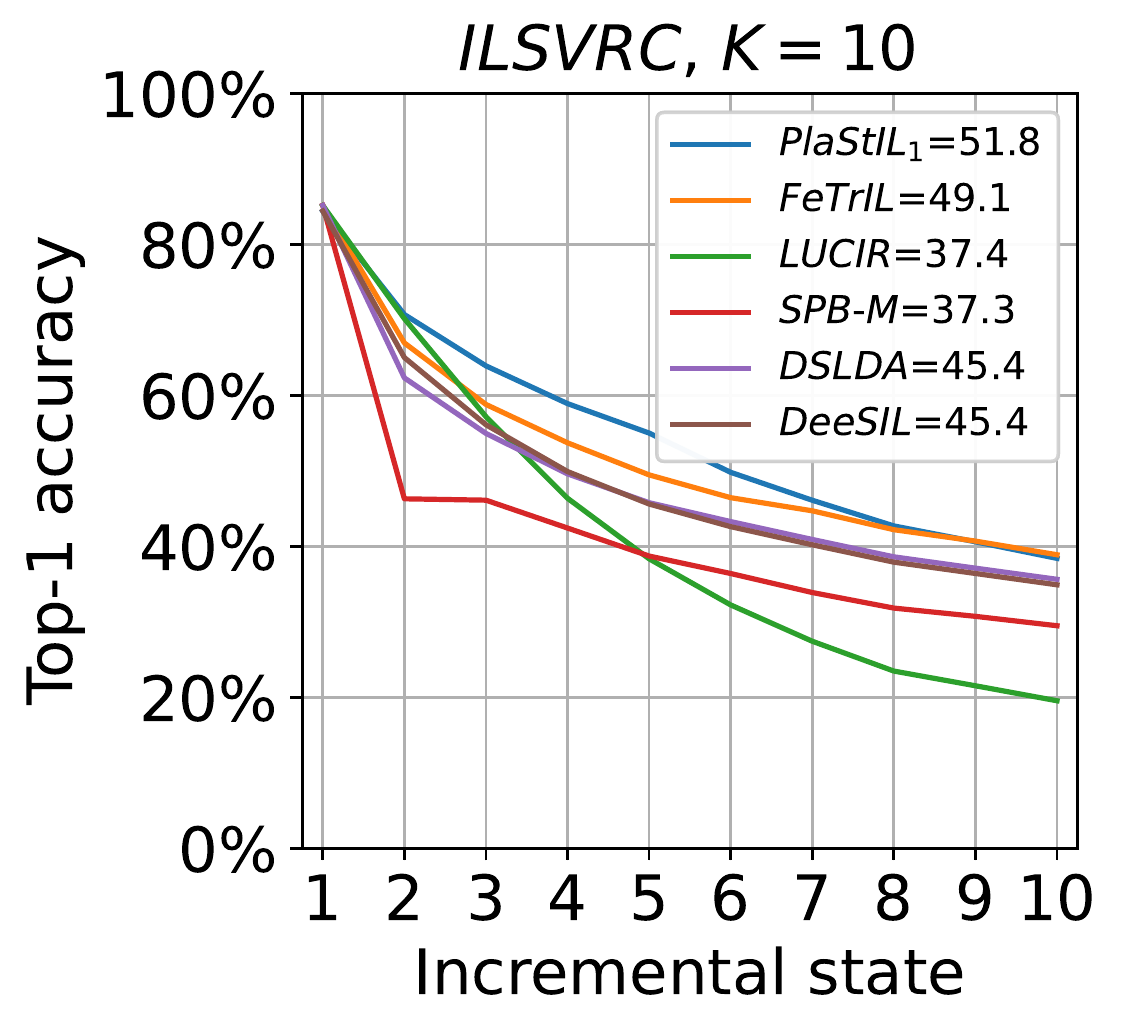} }}%
    {{\includegraphics[height=.3\linewidth]{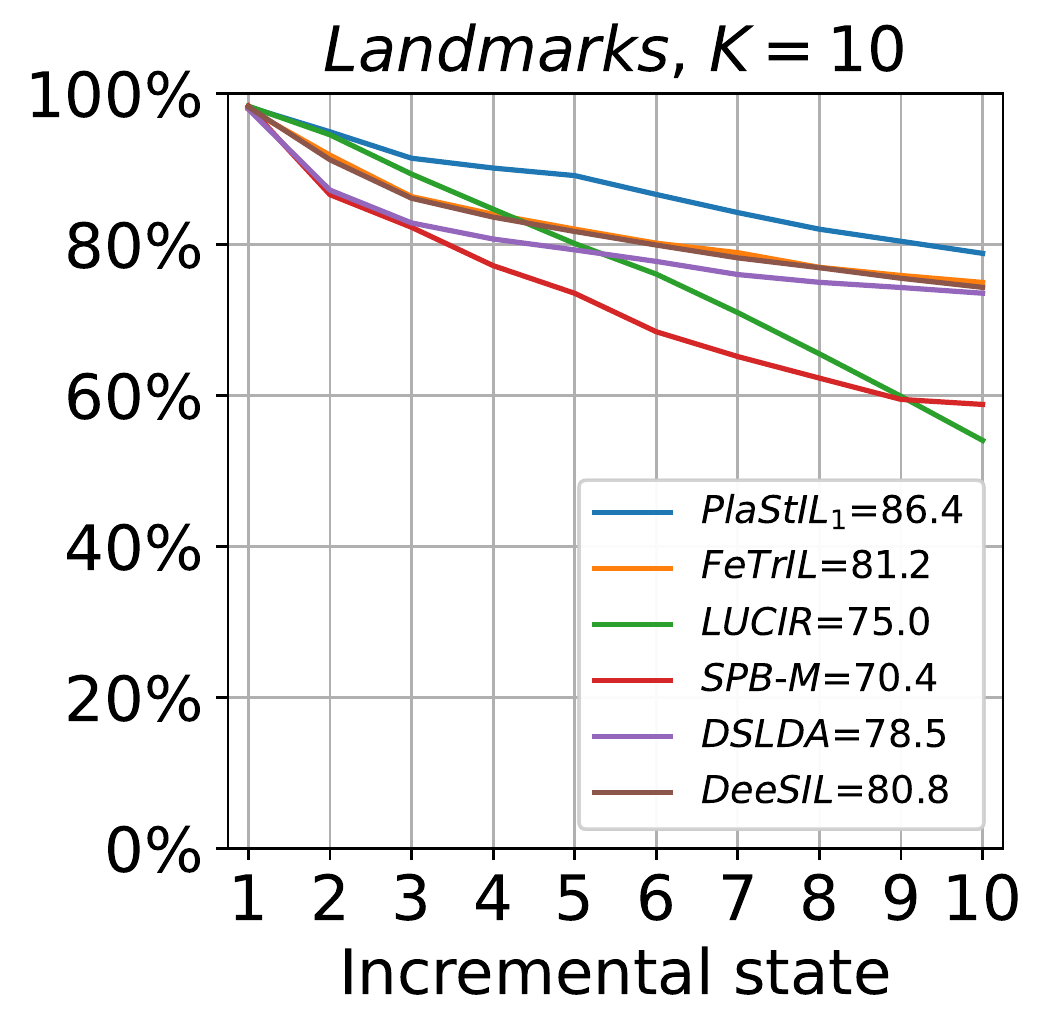} }}%
    {{\includegraphics[height=.3\linewidth]{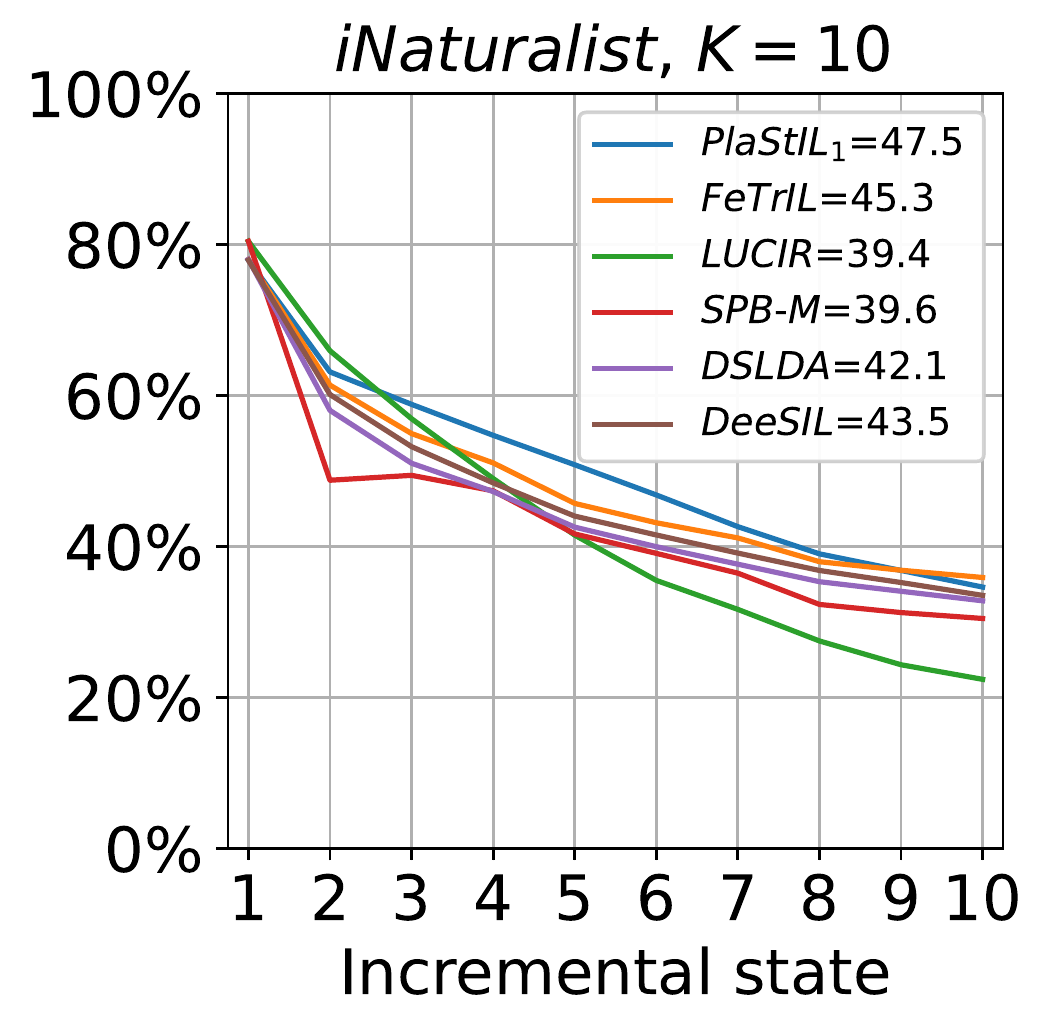} }}%

	\caption{Incremental accuracy  across all states for $K=10$. {Plots for for $K=5$ and $K=20$ are presented at Figure}~\ref{fig:complete_acc} {in the appendix.} Plots are presented for the best methods from Table~\ref{tab:main_results}. 
}
\label{fig:accuracy}
\end{figure*}

We propose a more detailed presentation of the incremental accuracy in Figure~\ref{fig:accuracy}.
These results confirm the large gap between the proposed methods and distillation-based ones.
\spbm has lower performance at the start of the process, but then becomes better than $LUCIR$, because of the representativeness of the initial model, as previously explained.

\subsection{Method analysis}
\label{subsec:analysis}

We conduct an analysis of \ourmodel in terms of model footprint and number of operations needed to infer test image predictions.
These experiments are run using \ourmodel applied on top of $DeeSIL$ since this variant has the best overall results.
They are important in order to highlight the merits and limitations of the proposed method.

\textbf{Model footprint.}
Incremental learning algorithms are particularly useful in memory-constrained environments.
Their model footprint is thus an important characteristic.
Distillation-based methods require the storage of models $\mathcal{M}_k$ and $\mathcal{M}_{k-1}$ to preserve past knowledge.
Transfer-based methods only need $\mathcal{M}_1$, but they optimize stability at the expense of plasticity.
The three versions of \ourmodel, whose performance is presented in Table~\ref{tab:main_results}, store $\mathcal{B}_1$, the initial model base and a variable number of model tops.
Each of the four recent model tops used by \ourmodelNospace$_1$ fine-tunes the last convolutional layer of ResNet-18~\citep{he2016_resnet}, which accounts for  21.45\% of total number of the model's parameters.
The parametric footprint of \ourmodelNospace$_1$ is thus lower than that of distillation-based methods.
\ourmodelNospace$_2$ has the same footprint as \ourmodelNospace$_1$ since it stores two tops which account for 42.9\% of ResNet-18 parameters each. 
As an ablation of \ourmodel, we also present results with a single model top, regardless of its fine-tuning depth \href{in Appendix}.
Naturally, \ourmodelNospace$_{all}$ obtains the best results in this configuration, but interesting gains are still obtained with model tops which fine-tune one or two final convolutional layers in \ourmodelNospace$_1$ and \ourmodelNospace$_2$, respectively. 


\begin{figure*}[t]
    \centering
    {{\includegraphics[height=.235\linewidth]{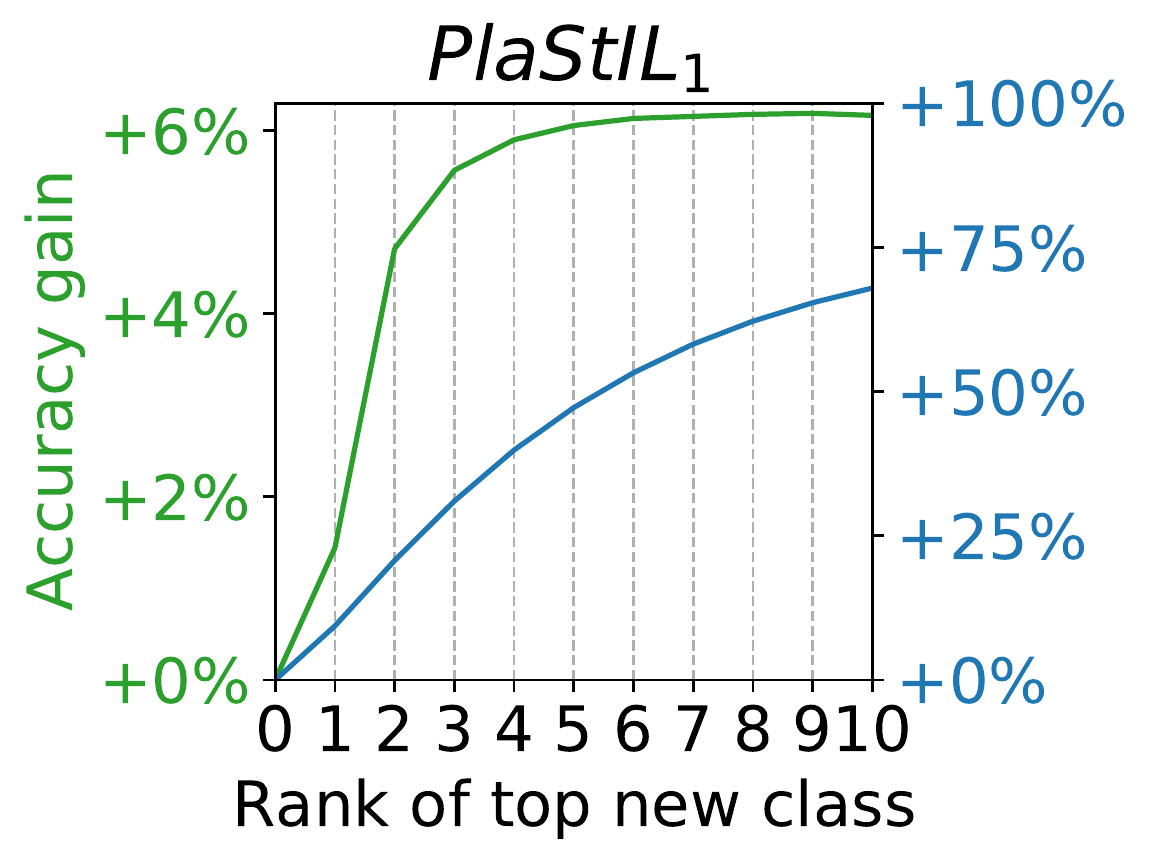} }}
    {{\includegraphics[height=.235\linewidth]{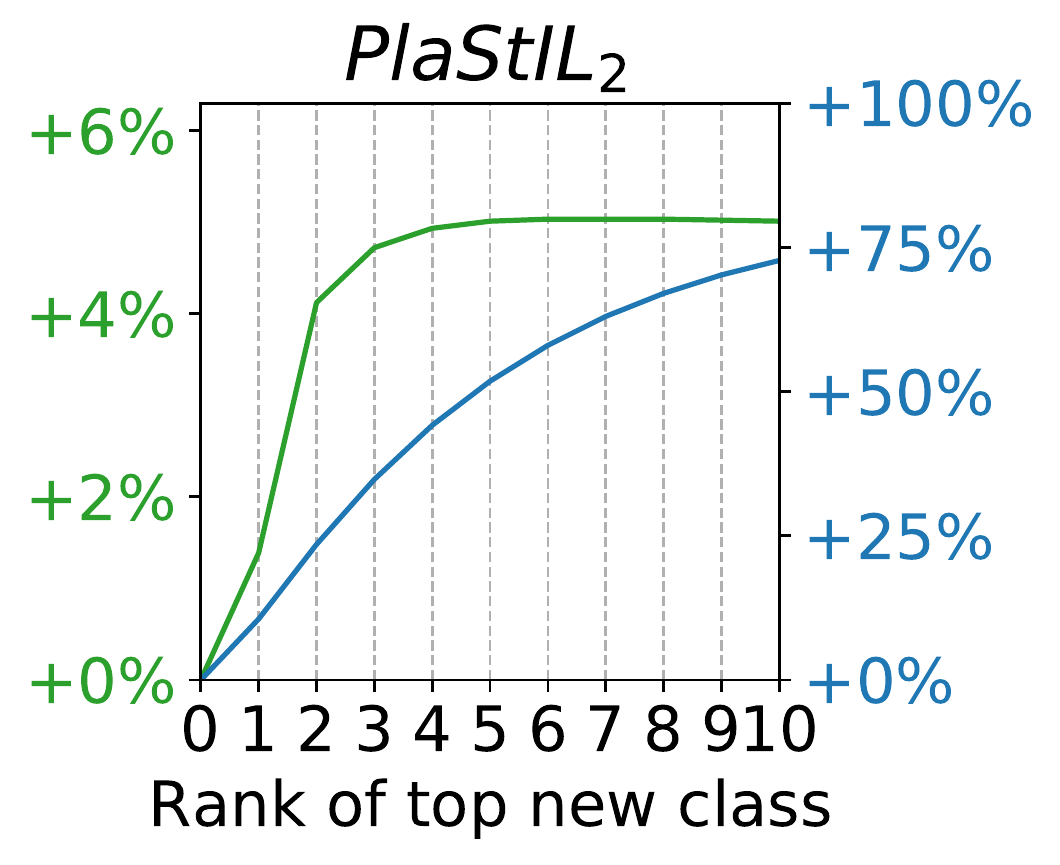} }}
    {{\includegraphics[height=.235\linewidth]{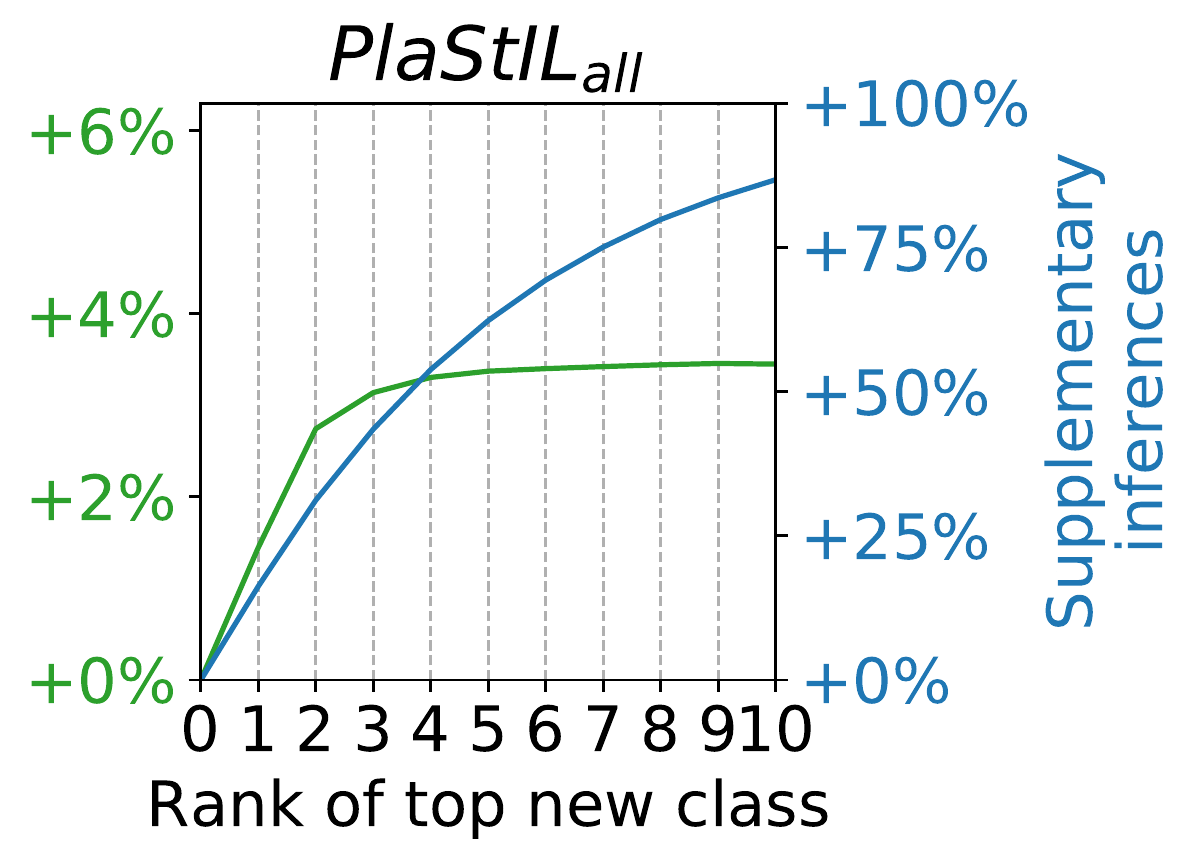} }}
    \caption{Top-1 accuracy gains obtained with the three variants of \ourmodel applied to $DeeSIL$ with different thresholds for the rank of the top new class among the predictions generated with Equation~\ref{eq:init}. Results are shown for $ILSVRC$ with $K=10$ states. The corresponding percentage of supplementary inferences needed for each threshold is also plotted. Interesting gains are obtained starting with a recent class predicted in the second position, which requires approximately 25\% of supplementary inferences for \ourmodelNospace$_{1}$. \textit{Best viewed in color.}}%
    \label{fig:inferences}%
\end{figure*}

\textbf{Inference complexity.}
The classification layer defined in Equation~\ref{eq:hybrid} is fed with features from the initial and the updated model top(s), for past and recent classes, respectively.
By default, \ourmodel inferences requires an extraction of features for all used model tops.
This supplementary inferences cost varies for \ourmodel variants due to the different number of parameters in the model top(s) that they use. 
{This supplementary computational cost can be reduced if predictions are first computed using the initial model only, as defined in Equation}~\ref{eq:init}. {Then, subsequent model tops are used only if one of their classes is strongly activated for the test image. A top is used for inferences only if at least one of its associated classes is ranked among the top classes of the of the classification layer from Equation}~\ref{eq:init}. 
{The closer to 1 this rank threshold is, the smaller the added computational cost will be since fewer tops are likely to be used for inference. However, a restriction to small ranks might also discard useful model tops and thus reduce the positive effect of} \ourmodel.
Evaluation is done with top ranks of new class in $\mathcal{W}_k^{fix}$ between 1 and 10 to examine the trade-off between inference complexity and performance.
The obtained accuracy, as well as the added inference cost, for the three variants of \ourmodel applied over $DeeSIL$ are presented in Figure~\ref{fig:inferences}.
The results show that performance gains relative to $DeeSIL$ raise sharply.
The best balance between performance gains and added costs is obtained for \ourmodelNospace$_{1}$.
This is explained by the fact that $\mathcal{T}$ have the lowest number of parameters for this variant.
Their activation can be done in a finer manner, resulting in a reduced overall inference cost. 
Interesting \ourmodel gains are obtained starting with a recent class being ranked second position by $DeeSIL$.
The accuracy curve becomes practically flat if a recent class is ranked beyond the third position by the baseline.
The results presented in Figure~\ref{fig:inferences} provide further support to the fact that \ourmodelNospace$_{1}$ is the most appropriate choice as plasticity layer added on top of $DeeSIL$.

\textbf{Choice of model tops.}
{The main experiments used model tops created for the most recent incremental states. Other top creation and removal policies are possible, and we compare the one proposed here with an oracle which performs an optimal selection of tops. The oracle selects model tops associated to different states so as to maximize the average incremental accuracy of the CIL process by aggregating the accuracy of different incremental states computed on the test set. The results from} Table~\ref{tab:model_top_choice} {indicate that the creation of tops for the most recent states provides a performance level which is close to the optimal one achievable by the oracle. This is probably explained by the fact that the evaluated CIL scenarios use a random assignment of classes to states. Other selection strategies might be more appropriate if the assumptions made about the order of arrival of classes or data were different, as discussed} in Subsection~\ref{subsec:method}.

\begin{table}
	\begin{center}
		\resizebox{0.98\textwidth}{!}{
			\begin{tabular}{@{\kern0.5em}lcccccccccccc@{\kern0.5em}}
				\toprule
				\multirow{2}{*}{\textbf{CIL Method}}
				& \multicolumn{3}{c}{\textit{ILSVRC}}
				&& \multicolumn{3}{c}{\textit{Landmarks}}
				&& \multicolumn{3}{c}{\textit{iNaturalist}}
				\\ \cmidrule(lr){2-4} \cmidrule(lr){6-8} \cmidrule(l){10-12}  
				
				& \multicolumn{1}{c}{$K$=5}
				& \multicolumn{1}{c}{$K$=10}
				& \multicolumn{1}{c}{$K$=20}
				&
				& \multicolumn{1}{c}{$K$=5}
				& \multicolumn{1}{c}{$K$=10}
				& \multicolumn{1}{c}{$K$=20}
				&
				& \multicolumn{1}{c}{$K$=5}
				& \multicolumn{1}{c}{$K$=10}
				& \multicolumn{1}{c}{$K$=20}
				
				\\ \midrule
				
				$DeeSIL$~\citep{belouadah2018_deesil}       & 52.4 & 45.4 & 37.5 && 87.4 & 80.8 & 73.8 && 52.7 & 43.5 & 33.9 \\  \hline 
				w/ \ourmodelNospace$_1$                    & 58.6 & 51.8 & 41.9 && 92.1 & 86.4 & 78.1 && 56.8 & 47.5 & 36.4 \\ 
				w/ \ourmodelNospace$_1$+oracle      & 58.6 & 51.8 & 42.1 && 92.1 & 86.4 & 78.7 && 56.8 & 47.8 & 36.8 \\  \hline
                w/ \ourmodelNospace$_2$                    & 59.2 & 50.2 & 39.7 && 92.2 & 85.1 & 76.7 && 57.3 & 46.9 & 36.0 \\ 
				w/ \ourmodelNospace$_2$+oracle      & 59.3 & 50.3 & 40.1 && 92.2 & 85.2 & 77.3 && 57.5 & 47.1 & 36.5 \\  \hline
				w/ \ourmodelNospace$_{all}$                & 57.7 & 48.6 & 39.0 && 90.7 & 83.4 & 75.3 && 58.2 & 47.1 & 35.6 \\ 
				w/ \ourmodelNospace$_{all}$+oracle   & 57.9 & 48.7 & 39.3 && 90.7 & 83.7 & 75.8 && 58.3 & 47.5 & 36.0 \\

				\toprule
			\end{tabular}
		}
	\end{center}
	\caption{Comparison of \ourmodel with a version that knowing the final composition of the different states will only fine-tune the relevant states (\ourmodel+oracle).
     ~The average gains are of +0.2\% with \ourmodel+oracle.}
	\label{tab:model_top_choice}
\end{table}

\section{Conclusion}
\label{sec:conclusion}

We proposed a new method which adds a plasticity layer to transfer-based methods in exemplar-free CIL.
Plasticity is improved by training dedicated model tops which fine-tune a variable number of deep models layers for one or more recent incremental states.
The predictions of the different model tops used by our method are integrated in a hybrid classification layer.
Model tops improve accuracy compared to the transfer-based method to which they are added to in order improve plasticity. 
These improvements are obtained by introducing supplementary parameters so as to have a total footprint which remains lower than that of distillation-based methods.
The comparison of these methods with \ourmodel is clearly favorable to the latter. 
The takeaway is that the parameters allocated to the previous model in distillation-based approaches are better spent on partially fine-tuned model tops. 
This finding is aligned with those reported in recent comparative studies which question the usefulness of distillation in large-scale CIL with or without memory~\citep{belouadah2021_study,masana2021_study,prabhu2020gdumb}. 
We believe that future studies should consider transfer-based methods to assess progress in exemplar-free CIL in a fair and comprehensive manner.
We plan to optimize the stability-plasticity compromise for EFCIL to further improve the encouraging results reported here.
First, we will try to devise a classifier which predicts the most probable initial state for each test sample, following the proposal made in~\citet{kumar2021_efficient}.
If successful, this prediction would reduce the classification space to individual states and remove the need for a hybrid classification layer.
Second, we will investigate the use of pretrained representations learned with larger amounts of supervised or semi-supervised data, as proposed in~\citet{dhamija2021self,hayes2020_remind,esn_2023}.
Such models can be seamlessly integrated in the \ourmodel pipeline.

\bibliography{collas2023_conference}
\bibliographystyle{collas2023_conference}

\appendix
\section{Appendix}

In this appendix, we provide: 
\begin{itemize}
\setlength\itemsep{0.1em}
    \item implementation details for all tested approaches; 
    \item details about datasets used in experiments;
    \item results for features transferability among datasets.
    \item {discussion on the different disk-memory usages}
    \item {additional detailed comparative accuracies}
\end{itemize}

\subsection{Implementation details}
As already mentioned in Section \ref{subsec:implementation}, we used the authors' optimal parameters to run all baselines. 
A ResNet-18 model \citep{he2016_resnet} and an $SGD$ optimizer with $momentum=0.9$ are used for all methods. We explicitly list the learning parameters of each method hereafter:

\subsubsection{Learning from scratch}
This type of learning is used to train models of the initial state (because it is not incremental), and also $Joint$, the upper bound method where all classes are learned with all their data at once.

Following \citet{belouadah2020_siw}, $Joint$ and the first models of $FT$ and $SIW$ \citep{belouadah2020_siw} are run for 120 epochs using $batch~size=256$ and $weight~decay=0.0001$. The $lr$ is set to 0.1 and is divided by 10 when the error plateaus for 10 epochs.

For $REMIND$ \citep{hayes2020_remind}, $Deep$-$SLDA$ \citep{hayes2020_deepslda}, $DeeSIL$ \citep{belouadah2018_deesil}, $Fixed_{NCM}$~\citep{rebuffi2017_icarl}, $FeTrIL$~\citep{petit2023fetril} and $PlaStIL$ (ours), we follow \citet{hayes2020_remind} and run the model for 90 epochs using $lr=0.1$, $batch~size=128$, $weight~decay=0.00001$. The $lr$ is set to its initial value decayed by 10 every 30 epochs. The $lr$ is constrained to do not decrease beneath 0.001.

For $LUCIR$, $LwF$, the first model is trained in the same manner than subsequent models (detailed below), following the authors of \citet{hou2019_lucir,rebuffi2017_icarl}.

\subsubsection{Incremental Learning}
Here, we describe the hyper-parameters used to train the models incrementally for model-update-based methods.

\begin{itemize}[noitemsep]
    \item $FT$~\citep{belouadah2020_siw} -  IL models are trained for 35 epochs with $batch~size=256$, $momentum=0.9$ and $weight~decay=0.0001$.  The learning rate is set to $lr=0.1 / t$ at the beginning of each incremental state $(t\ge2)$ and is divided by 10 when the error plateaus for 5 consecutive epochs.
    
    \item $LwF$~\citep{rebuffi2017_icarl} - all models are trained for 60 epochs using $lr=1.0$, $batch~size=128$, and $weight~decay=0.0001$. The learning rate is divided by 5 at epochs 20, 30, 40, and 50.
    
    \item $LUCIR$~\citep{hou2019_lucir}  - all models are trained for 90 epochs using $lr=0.1$, $batch~size=128$ and $weight~decay=0.0001$. The $lr$ is divided by 10 at epochs 30 and 60. The method-specific parameters are the same as those from the original paper \citep{hou2019_lucir} and can also be found once we release the codes and configuration files. 
        
    \item $SIW$~\citep{belouadah2020_siw}  - is trained using the same hyper-parameters of $FT$ following the authors.

    \item $SPB$-$M$~\citep{wu2021striking}  - all models are trained for 90 epochs using $lr=0.1$, $batch~size=128$ and $weight~decay=0.0001$. The $lr$ is divided by 10 at epochs 30 and 60. The method-specific parameters are the same as those from the original paper \citep{wu2021striking} and can also be found once we release the codes and configuration files. Note that $SPB$-$M$~\citep{wu2021striking} has no available code, we implemented it by modifying significantly the code of $LUCIR$~\citep{hou2019_lucir}, and verified its correctness on the results the authors provided in their paper.
    
    \item {$REMIND$}~\citep{hayes2020_remind}  {- method-specific parameters are the same as those from the original paper, and run from the available code, using the advised version of pytorch}
\end{itemize}
We reimplemented the method, and verified its correct functioning using the protocol from the original paper. If the paper is accepted we will publish the code of every implementation we wrote in order to facilitate reproducibility.

\subsection{Datasets details}
\begin{table*}[t]
	\begin{center}
		\resizebox{\linewidth}{!}{
			\begin{tabular}{lcccccccccccc}
			\hline 
				
				Init. dataset & \multicolumn{1}{c}{\small $Landmarks$}&& \multicolumn{1}{c}{\small $iNaturalist$}&  \multicolumn{1}{c}{\small $ILSVRC$}  && \multicolumn{1}{c}{\small $iNaturalist$}&  \multicolumn{1}{c}{\small $Landmarks$} && \multicolumn{1}{c}{\small $ILSVRC$} \\

				\cmidrule(lr){2-4} 	\cmidrule(lr){5-7} \cmidrule(lr){8-10}
				IL dataset &  \multicolumn{3}{c}{\small $ILSVRC$}&  \multicolumn{3}{c}{\small $Landmarks$}&  \multicolumn{3}{c}{\small $iNaturalist$}  \\
			\hline 
				
				\small $Deep$-$SLDA$~\citep{hayes2020_deepslda} & 19.3 && 29.2 &  67.8 && 60.1 & 16.1 &&  35.2 &  \\
				\small $DeeSIL$~\citep{belouadah2018_deesil}  & 21.6  &&  29.6  &  70.5 &&  61.9 & 16.2 &&  36.1 &  \\
				\small $DeeSIL$ w/ \ourmodelNospace$_1$   & \textbf{28.1} && \textbf{33.8} &  \textbf{76.8} && \textbf{68.3} & \textbf{21.4} && \textbf{39.2} &  \\
				
			\hline 
				\multicolumn{1}{l}{$Joint$} &
				\multicolumn{3}{c}{72.98} & 
				\multicolumn{3}{c}{97.41}  &
				\multicolumn{3}{c}{75.60} \\
			\hline
			\end{tabular}
		}
	\end{center}

	\caption{Average top-1 incremental accuracy  in a dataset transfer learning configuration. Results are given for transferring between all pairs of initial and target datasets. All experiments are run with $K=10$ states.
		\textbf{Best results are in bold}. 
	}
	\label{tab:transfer_results}
\end{table*}

\begin{table}[]
	\begin{center}
		\resizebox{0.7\textwidth}{!}{
			\begin{tabular}{|l|c|c|c|c|c|}
			\hline
	Dataset & \#Train &  \#Test & $\mu$(Train) & $\sigma$(Train)\\	
			\hline
   	$ILSVRC$~\citep{olga2015_ilsvrc} &  1,231,167 & 50,000  & 1231.2 & 70.2\\		
                \hline                
    $Landmarks$~\citep{noh2017_landmarks} & 374,367 & 20,000  & 374.4 & 103.8\\		
                \hline    
                
    $iNaturalist$~\citep{van2018inaturalist} & 300,00 & 10,000 & 300.0 & 0.0 \\		
                \hline   
			\end{tabular}
		}
	\end{center}
	\caption{Summary of datasets. $\mu$ is the mean number of train images per class and $\sigma$ is the standard deviation}
	\label{tab:datasets}
\end{table}

The datasets used in evaluation are designed for three visual classification tasks: object, natural species and landmark recognition. Their main statistics are in tables~\ref{tab:datasets}.

\subsection{Feature transferability.}
The results from Table~\ref{tab:main_results} show that transfer-based methods work well when features are transferred within the same dataset.
In Table~\ref{tab:transfer_results}, we examine the behavior of \ourmodel in a transfer scenario which involves a domain gap.
The transfer is done between all pairs of initial and incremental datasets.
The results show that \ourmodelNospace$_1$ is more resilient to transfer compared to the transfer-based baselines.
The best results are obtained with $ILSVRC$ as initial models and $iNaturalist$ and $Landmarks$ as incremental datasets.
This is intuitive since $ILSVRC$ contains more diversified concepts and produces more generic features. 
It also has more samples per class compared to the other two datasets and its feature extractor is more robust.
The accuracy obtained with  a transfer from $ILSVRC$ is comparable with that of best distillation-based methods from Table~\ref{tab:main_results}.
The results with $iNaturalist$ as initial dataset are lower, but still interesting. 
Performance is lower when the domain shift between the initial and the target datasets is more important and if the initial model is learned on domain-specific data.
This is, for instance, the case when $Landmarks$ is used to train the initial model since this dataset only includes geographic landmarks. 
Note that transfer would be more efficient if larger initial datasets were used to reinforce the initial representation, as proposed in~\citet{belouadah2018_deesil,hayes2020_deepslda}.
However, for fairness, our focus is on transfer experiments which use the same number of initial classes as in Table~\ref{tab:main_results}.

{\subsection{Disk-memory usage.}}

\begin{figure*}[t]

	\centering
	\includegraphics[width=0.242\linewidth]{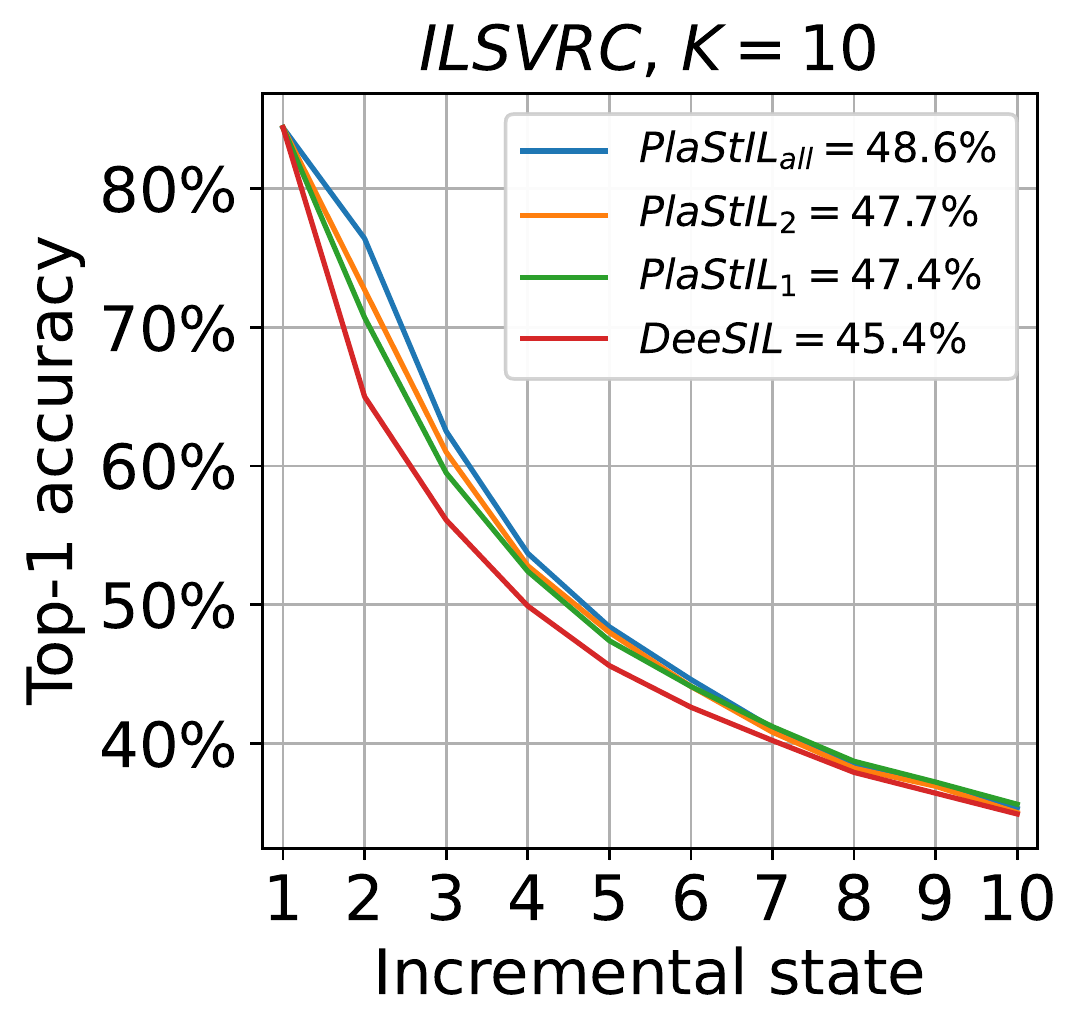}
	\includegraphics[width=0.227\linewidth]{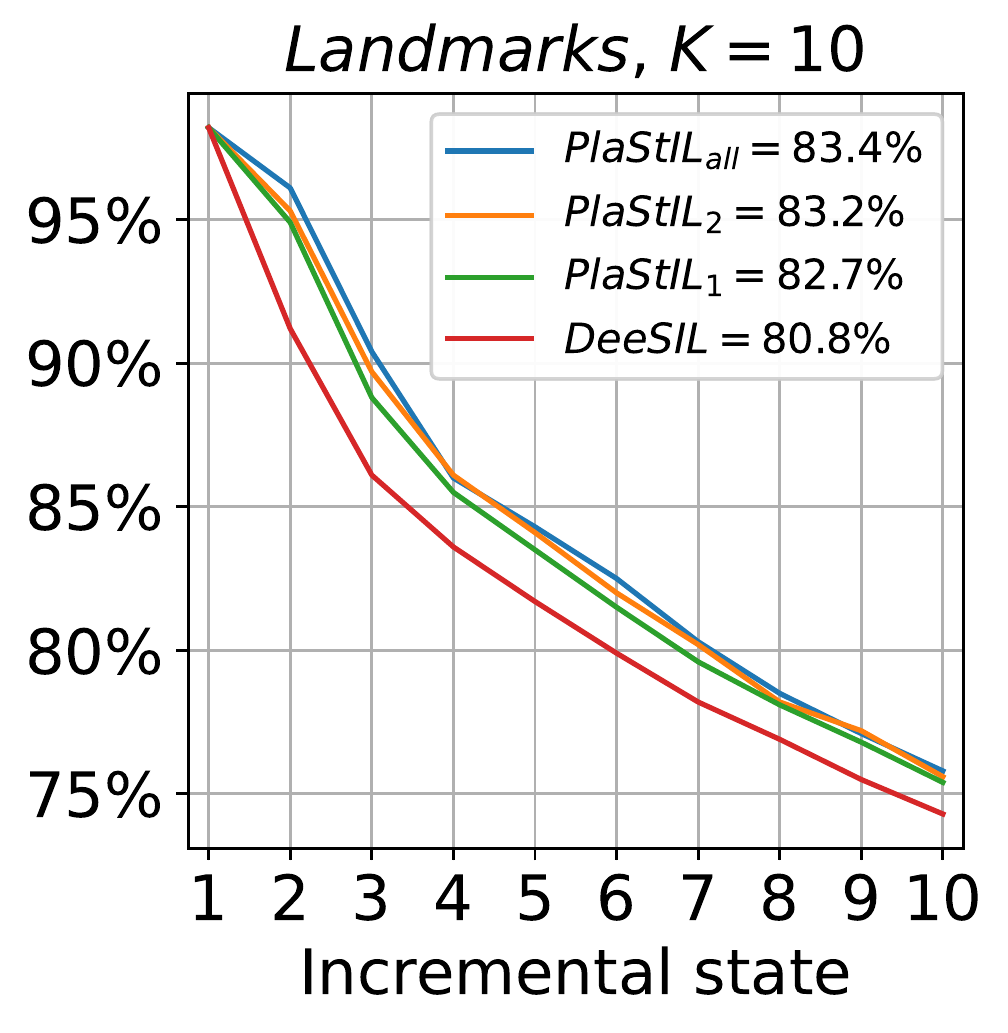}
	\includegraphics[width=0.227\linewidth]{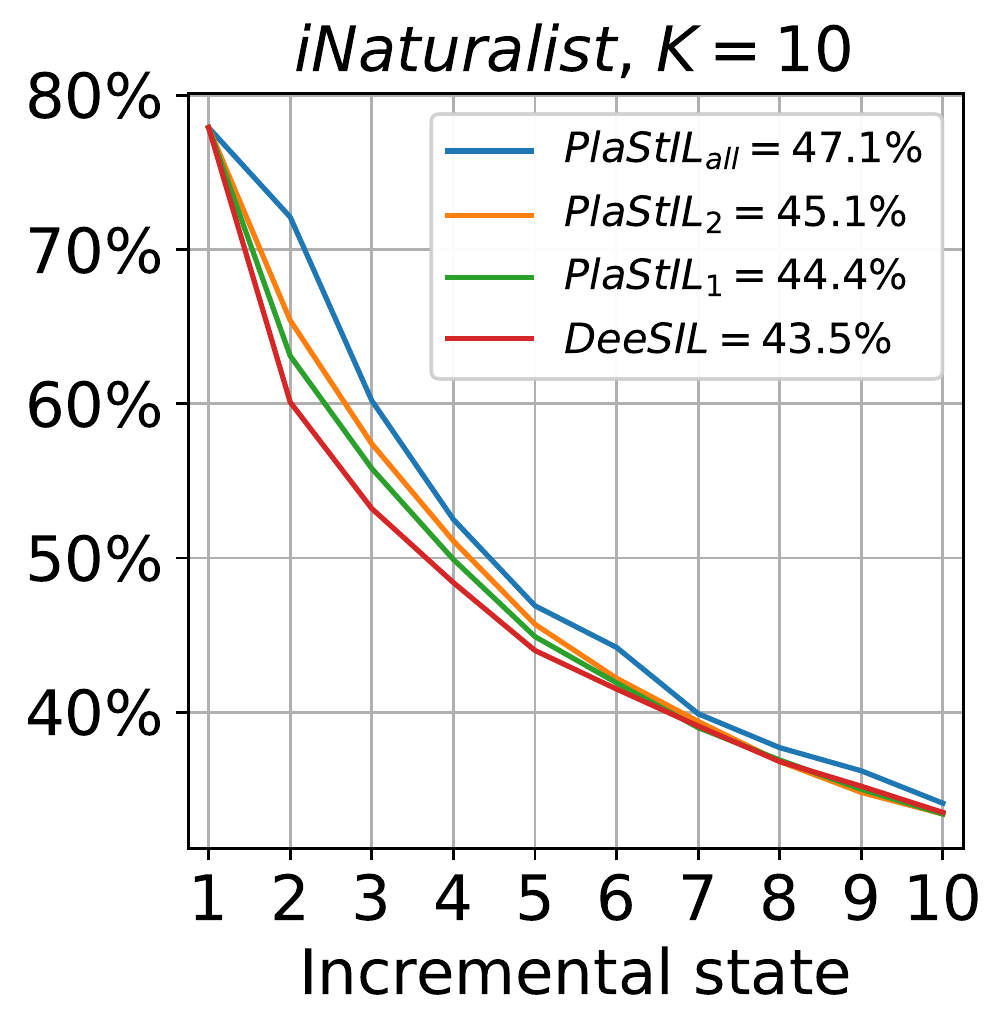}
	\caption{Top-1 incremental accuracy of three versions of \ourmodel applied to $DeeSIL$ when using a single model top with variable fine-tuning depth. $DeeSIL$ is a limit case in which the whole feature extractor is frozen. \textit{Best viewed in color.}}
	\label{fig:footprint}
\end{figure*}

\begin{figure*}[t]

	\centering
	\includegraphics[height=0.25\linewidth]{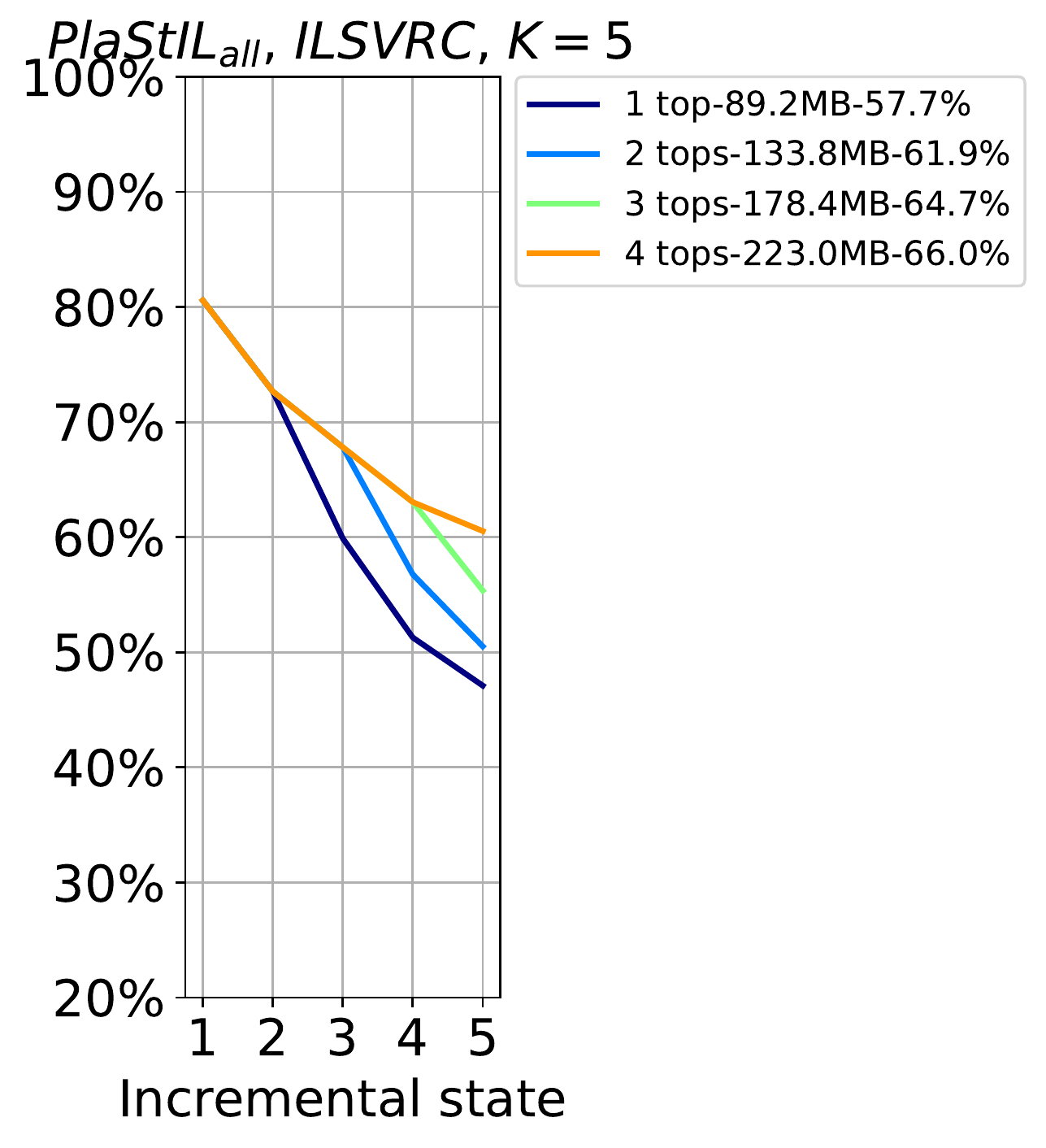}
	\includegraphics[height=0.25\linewidth]{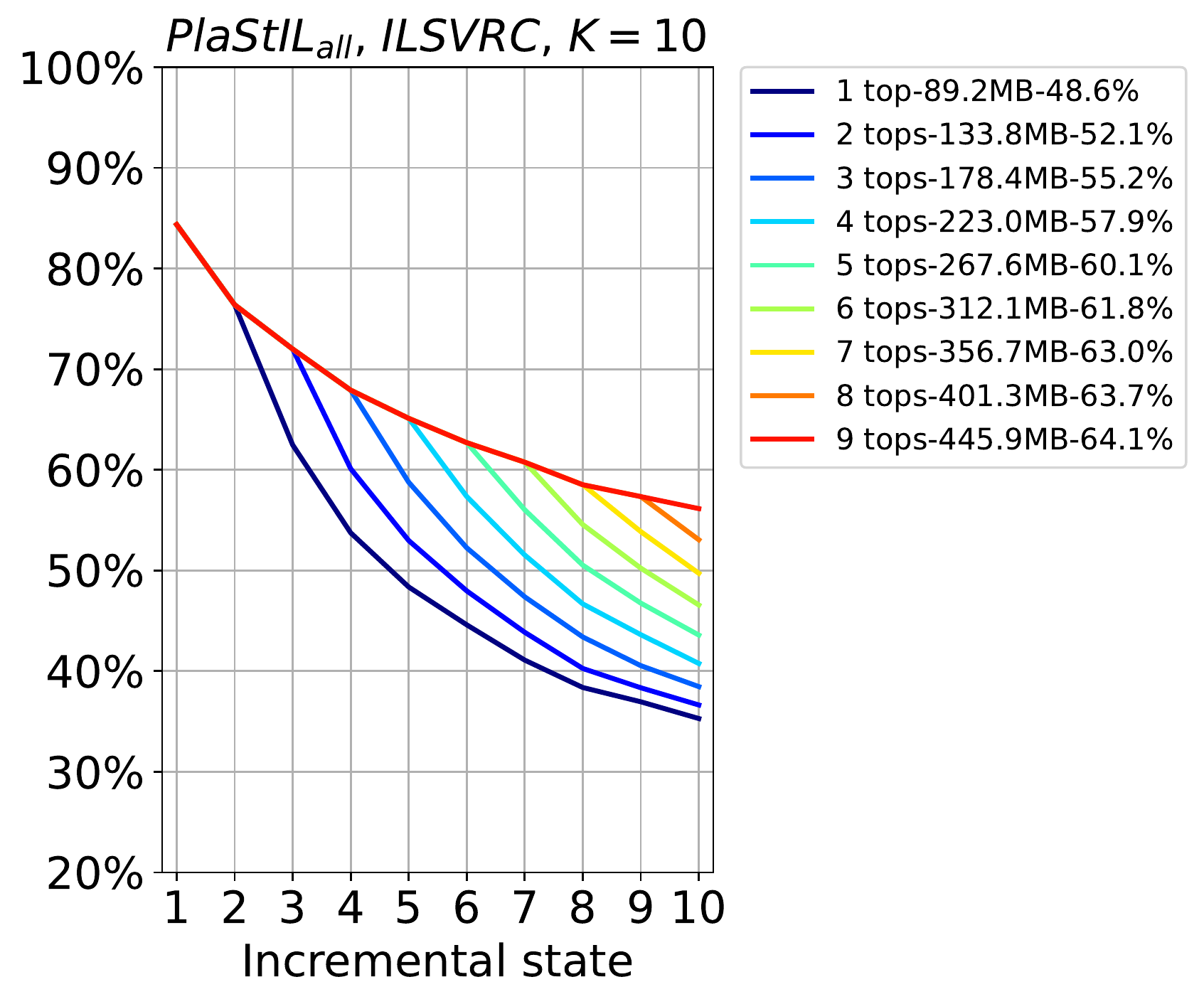}
	\includegraphics[height=0.25\linewidth]{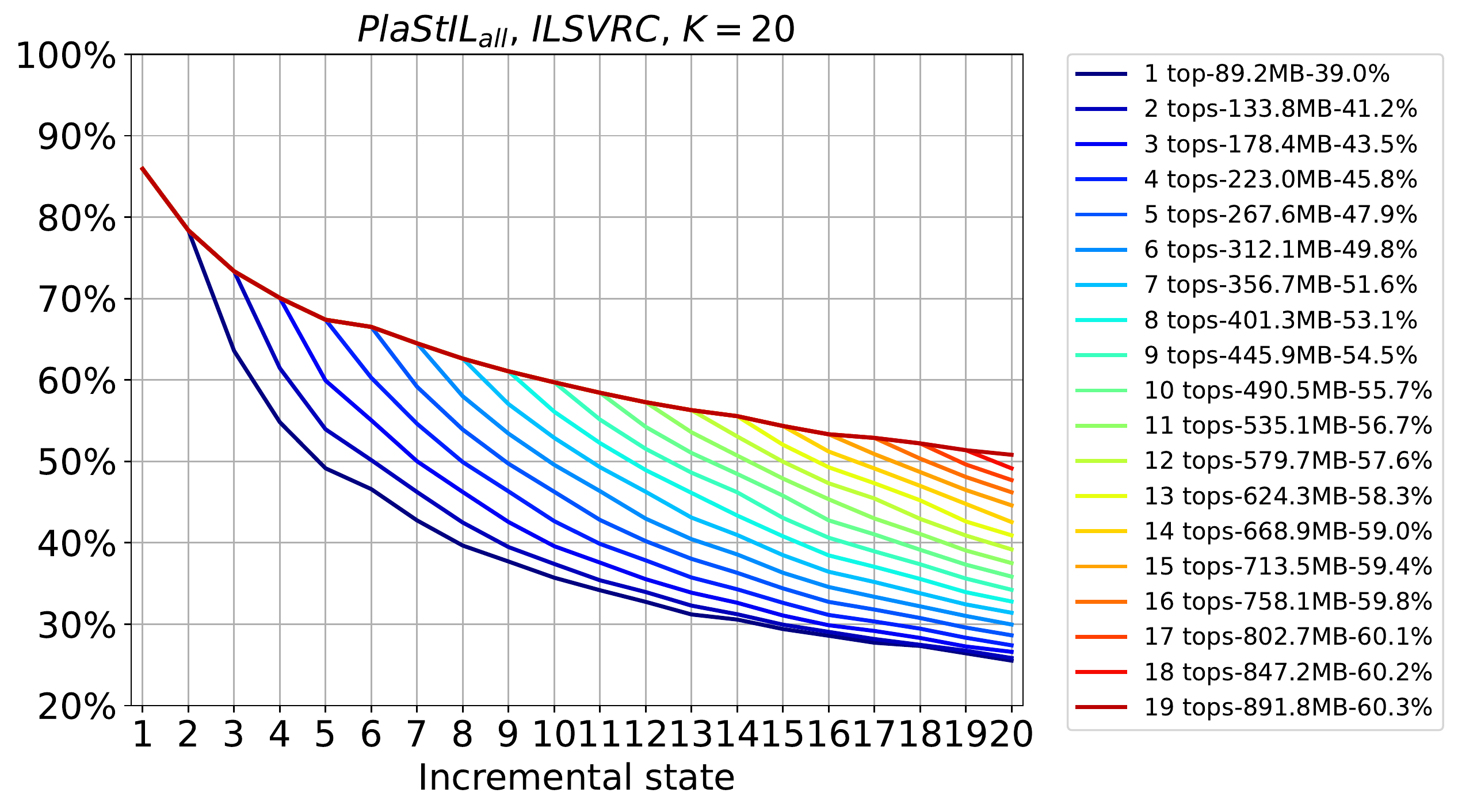}
	\includegraphics[height=0.25\linewidth]{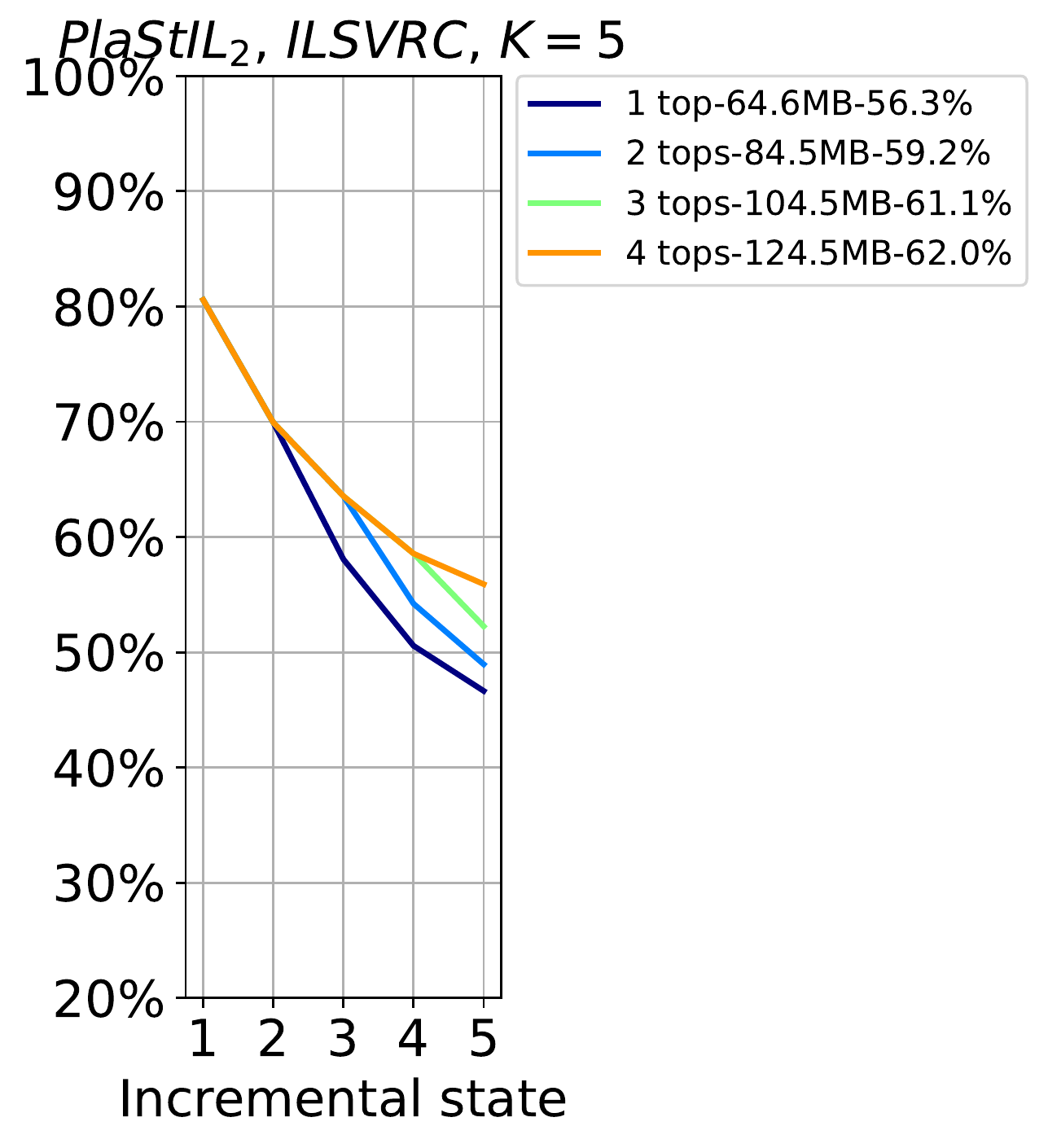}
	\includegraphics[height=0.25\linewidth]{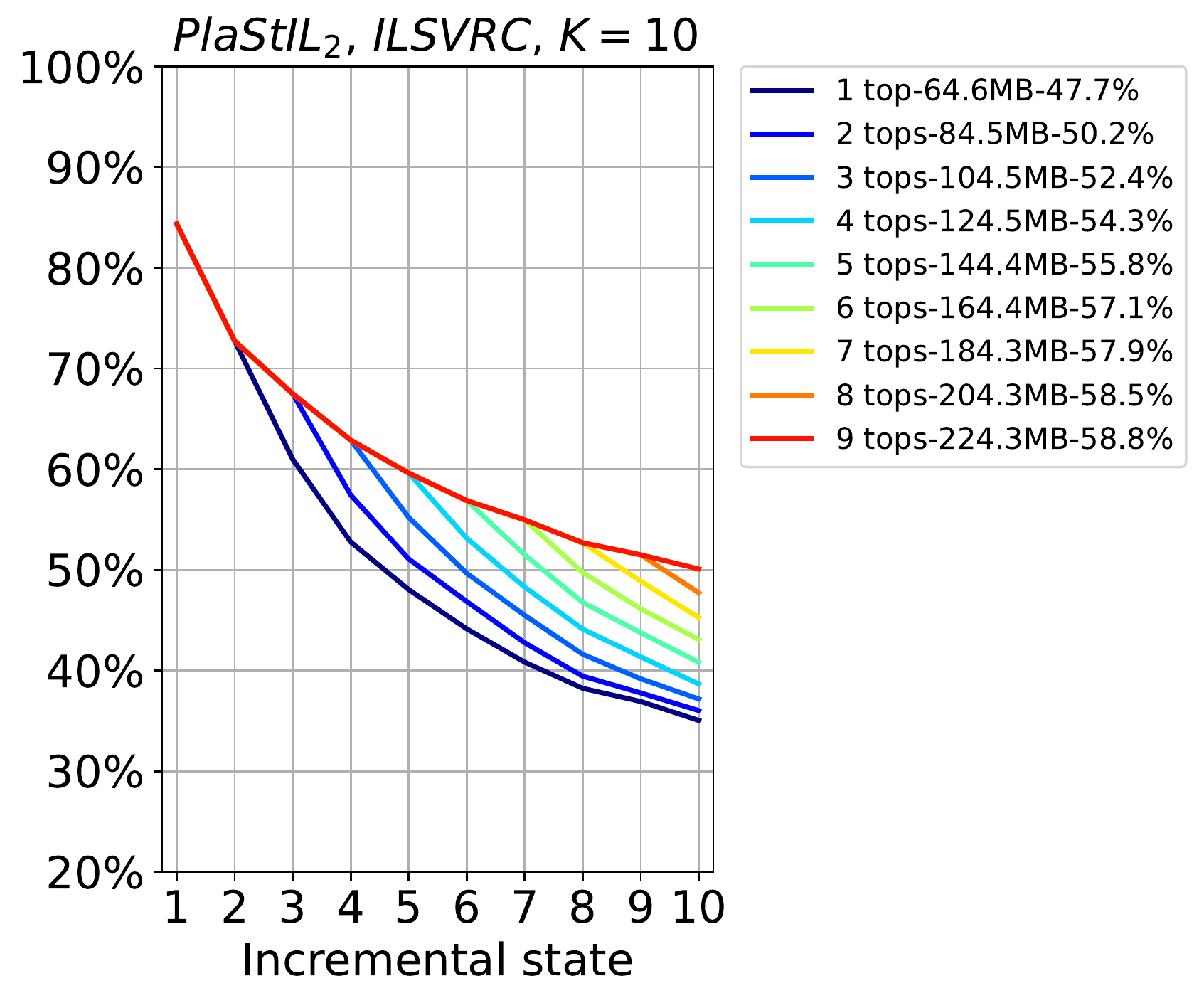}
	\includegraphics[height=0.25\linewidth]{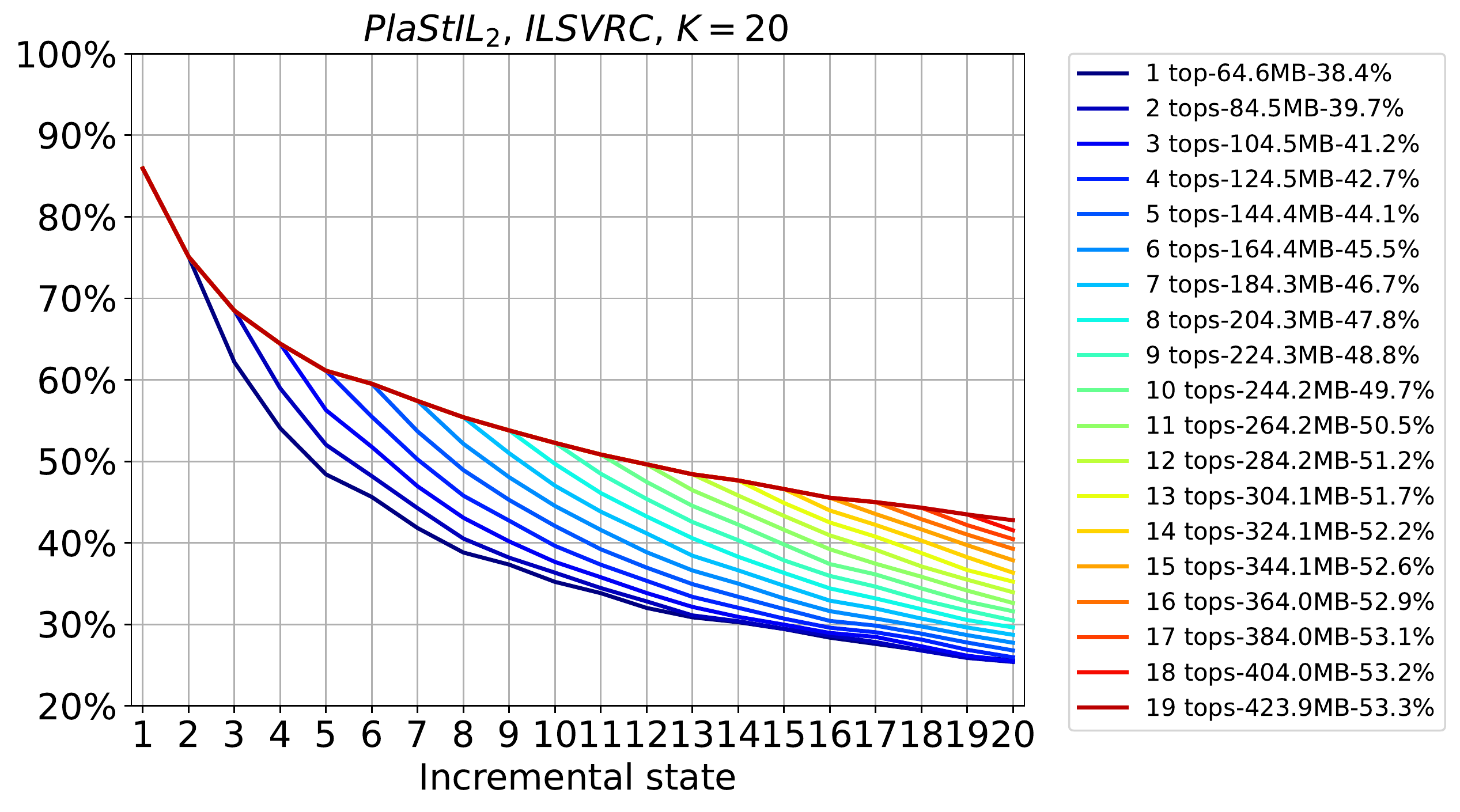}
	\includegraphics[height=0.25\linewidth]{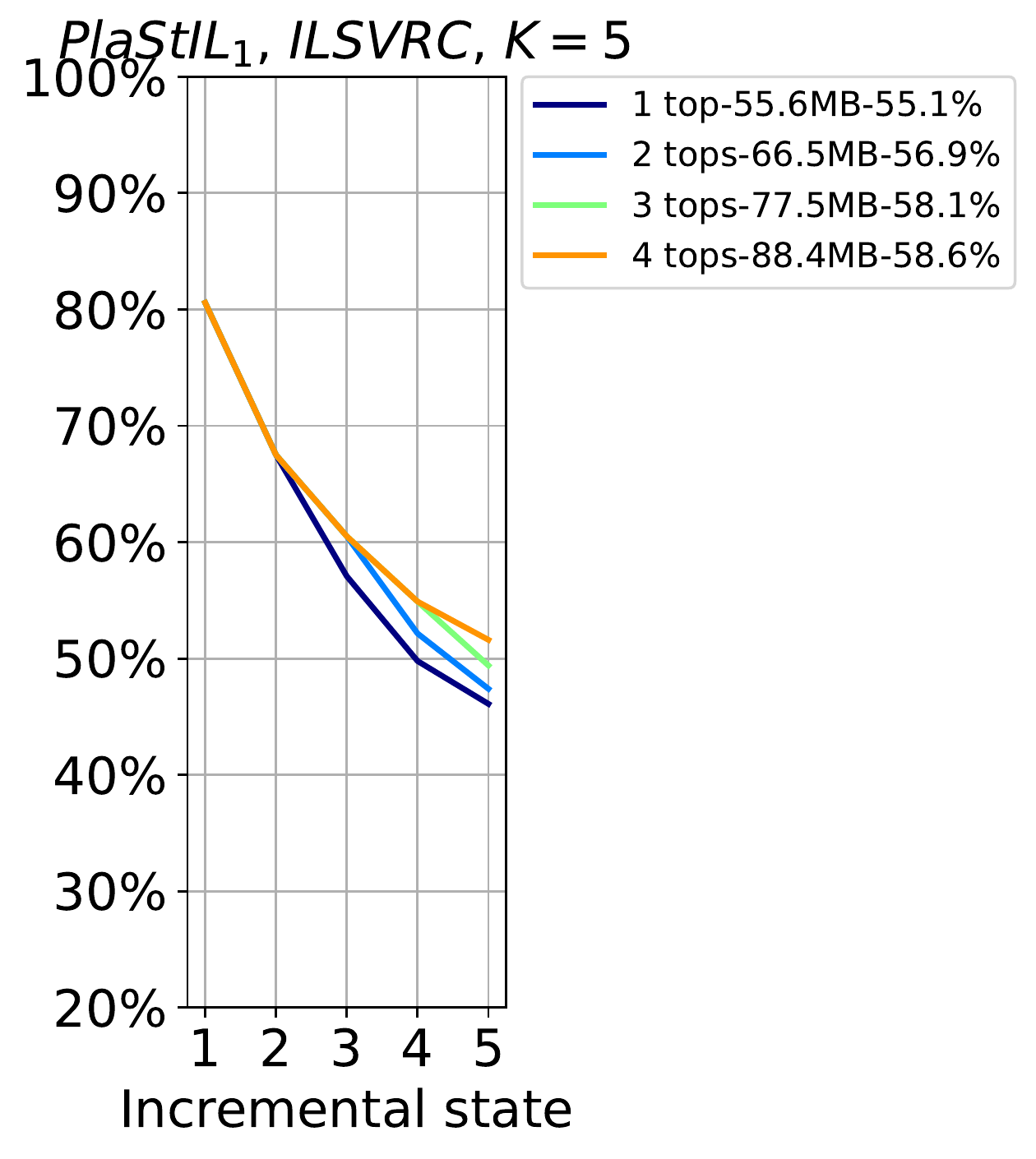}
	\includegraphics[height=0.25\linewidth]{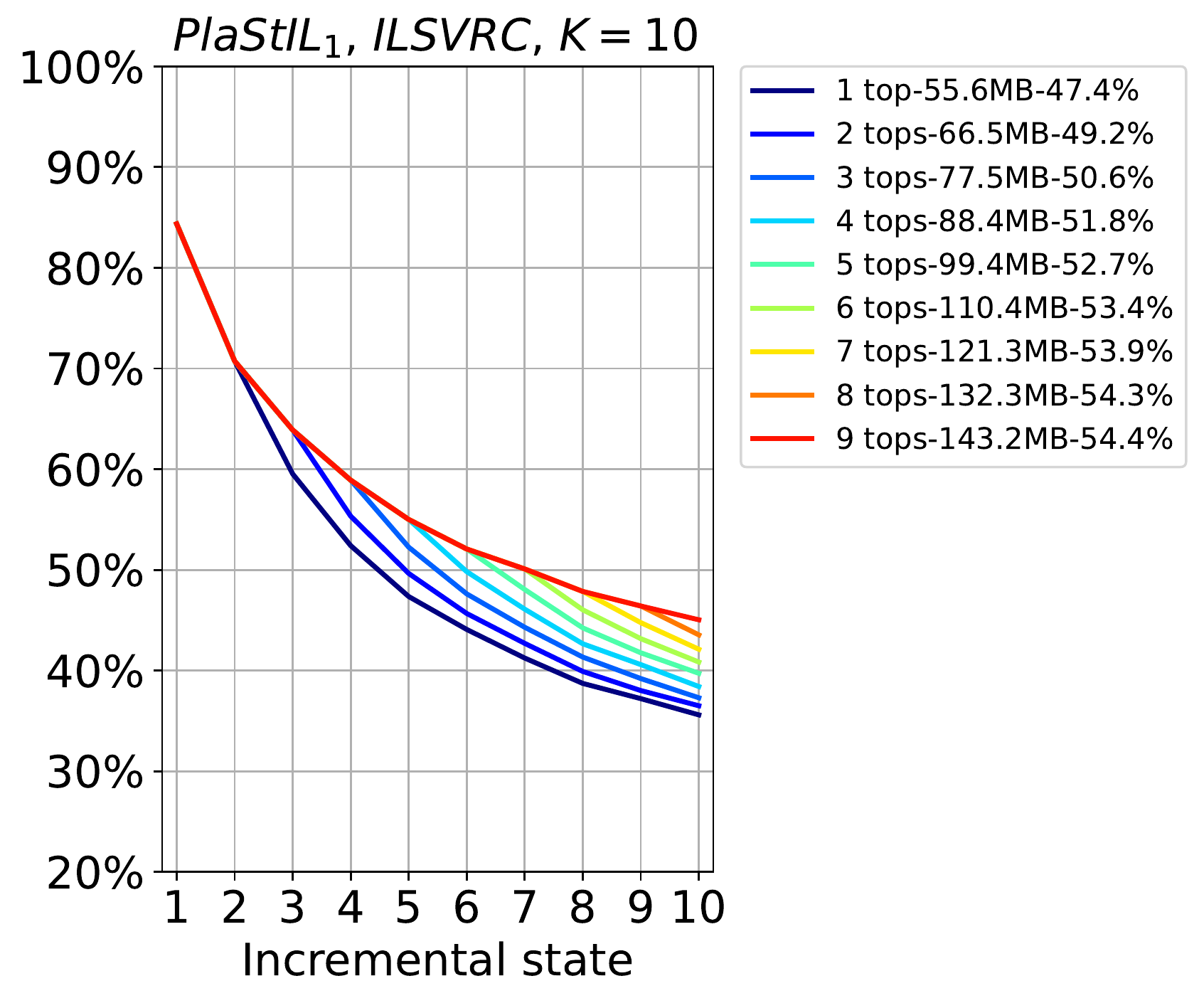}
	\includegraphics[height=0.25\linewidth]{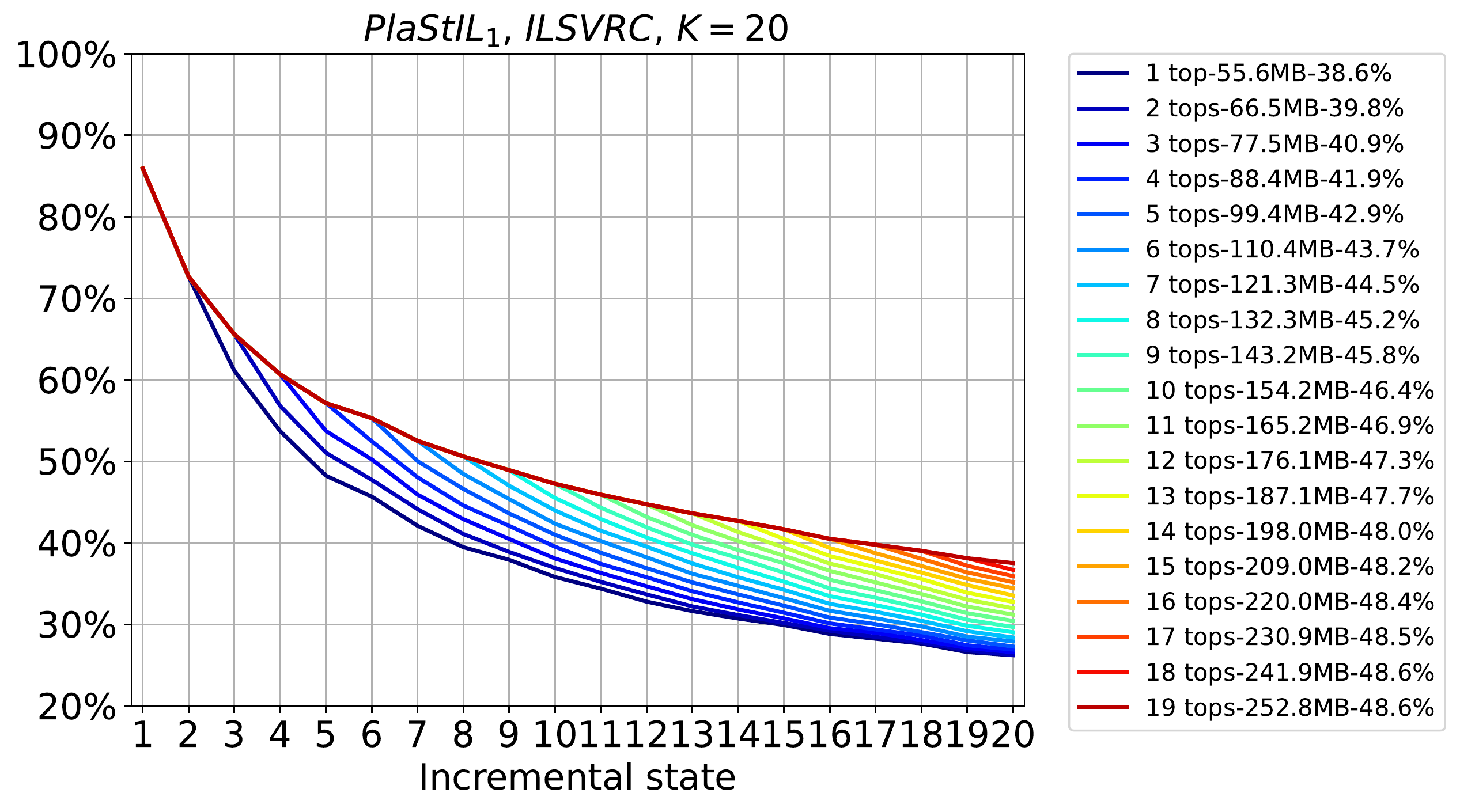}
 
	\caption{{Effect of varying the number of additional tops on incremental accuracy, on $ILSVRC$ with $K\in\{5,10,20\}$. As the number of tops increases, the accuracy of the model improves. However, this comes at the cost of increased memory usage on disk. \textit{Best viewed in color.}}}
	\label{fig:ailesderaie}
\end{figure*}
{The results of Figure}~\ref{fig:ailesderaie}  {demonstrate that as the number of tops increases, the accuracy of the model improves. However, it is important to note that this improvement in accuracy comes at the cost of increased memory usage on disk. This trade-off between accuracy and resource usage is a crucial consideration for optimizing model performance in practice in an exemplar-free setting.}

\begin{table}[h]
	\begin{center}
		\resizebox{0.85\textwidth}{!}{
			\begin{tabular}{@{\kern0.5em}llcccccccccccc@{\kern0.5em}}
				\toprule
				\multirow{2}{*}{\textbf{CIL Method}}
				& \multirow{2}{*}{\textbf{{mem on disk}}}
				& \multicolumn{3}{c}{\textit{ILSVRC}}
				&& \multicolumn{3}{c}{\textit{Landmarks}}
				&& \multicolumn{3}{c}{\textit{iNaturalist}}
				\\ \cmidrule(lr){3-5} \cmidrule(lr){7-9} \cmidrule(l){11-13}  
				&
				& \multicolumn{1}{c}{$K$=5}
				& \multicolumn{1}{c}{$K$=10}
				& \multicolumn{1}{c}{$K$=20}
				&
				& \multicolumn{1}{c}{$K$=5}
				& \multicolumn{1}{c}{$K$=10}
				& \multicolumn{1}{c}{$K$=20}
				&
				& \multicolumn{1}{c}{$K$=5}
				& \multicolumn{1}{c}{$K$=10}
				& \multicolumn{1}{c}{$K$=20}
				
				\\ \midrule
				$REMIND$~\citep{hayes2020_remind} \tiny(ECCV'20) & {89.18MB} & {52.2} & {44.8} & {35.9} && {83.3} & {77.5} & {72.2} && {50.6} & {39.4} & {31.3} \\
				$REMIND$~\citep{hayes2020_remind} \tiny(ECCV'20) & {133.78MB} & {52.3} & {44.9} & {36.2} && {83.4} & {79.3} & {75.8} && {50.9} & {43.8} & {35.0} \\
				$REMIND$~\citep{hayes2020_remind} \tiny(ECCV'20) & {222.96MB} & {52.3} & {44.4} & {\textbf{39.7}} && {84.2} & {81.0} & \underline{{78.0}} && {53.7} & {46.5} & {\textbf{37.8}} \\
                \hline
				$DeeSIL$~\citep{belouadah2018_deesil}& {44.59MB}& 52.4 & 45.4 & 37.5 && 87.4 & 80.8 & 73.8 && 52.7 & 43.5 & 33.9 \\ 
				w/ \ourmodelNospace$_1$            & {88.44MB} & \underline{58.6} & \textbf{51.8} & 41.9 && \underline{92.1} & \textbf{86.4} & \textbf{78.1} && 56.8 & \textbf{47.5} & \underline{36.4} \\ 

                w/ \ourmodelNospace$_2$       & {84.52MB} & \textbf{59.2} & \underline{50.2} & \textbf{39.7} && \textbf{92.2} & \underline{85.1} & 76.7 && \underline{57.3} & 46.9 & 36.0 \\ 
				w/ \ourmodelNospace$_{all}$      & {89.18MB} & 57.7 & 48.6 & \underline{39.0} && 90.7 & 83.4 & 75.3 && \textbf{58.2} & \underline{47.1} & 35.6 \\ 

				\toprule
			\end{tabular}
		}
	\end{center}
	\caption{{Average top-1 accuracy with three numbers of states} $K$ {per dataset, comparison with} $REMIND$ {with different budgets for the storage of their compressed vectors (1, 2 and 4 times the size of a ResNet18 on disk)}.  \textbf{Best results - in bold}, \underline{second best - underlined}.}
	\label{tab:appendix_results_vs_remind}
\end{table}

{Results from Table}~\ref{tab:appendix_results_vs_remind} {show that}  \ourmodel {strategy of storing model tops suits better the experiments than storing compressed representation vectors, even for a larger memory on disk.}
\newpage
{\subsection{Detailed comparative accuracies.}}

{In this section, we presented the incremental accuracy across all states for $K=5$ and $K=20$. These results confirm the large gap between the proposed methods and distillation-based ones.}
\begin{figure*}[h]

	\centering
	\includegraphics[height=0.25\linewidth]{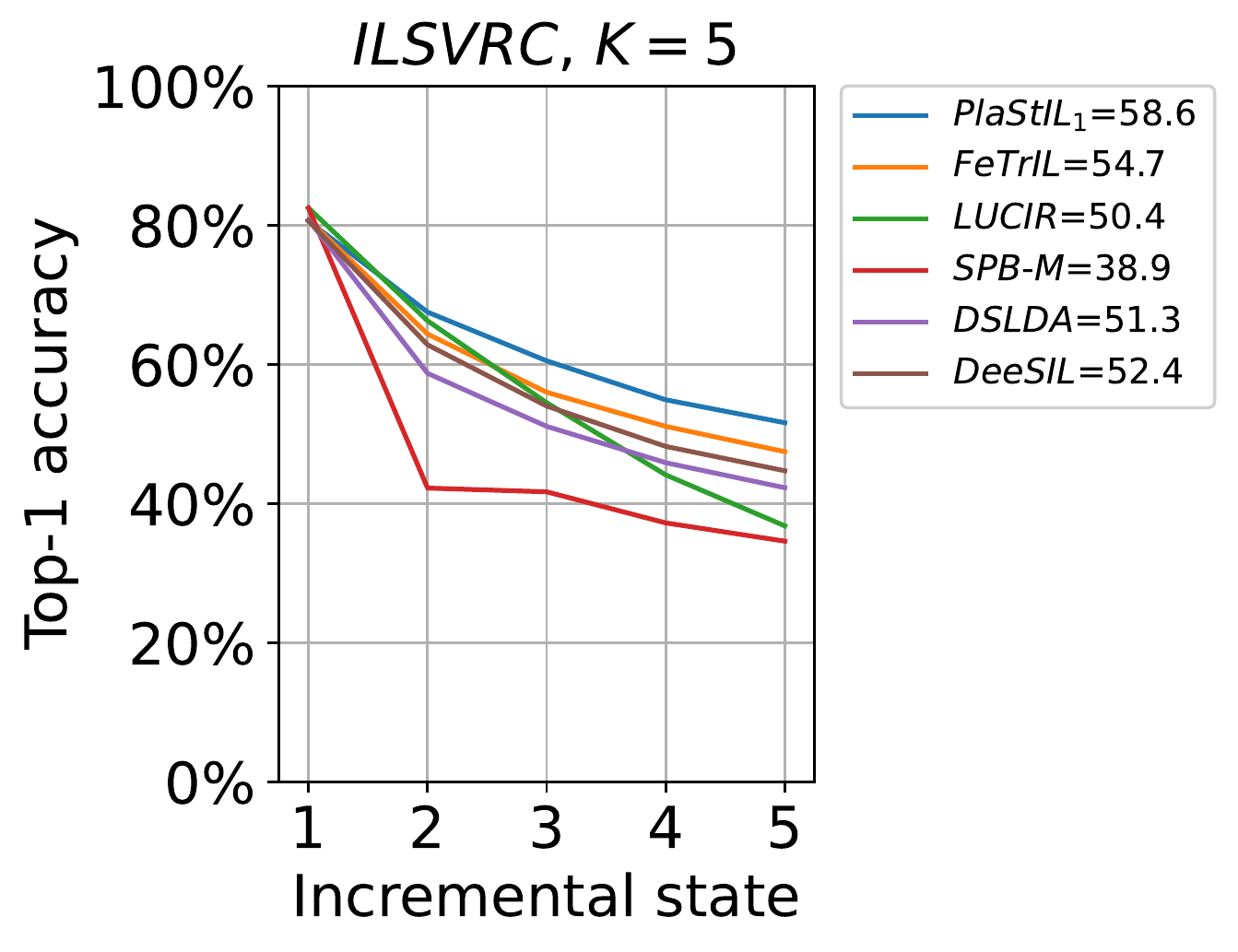}
	\includegraphics[height=0.25\linewidth]{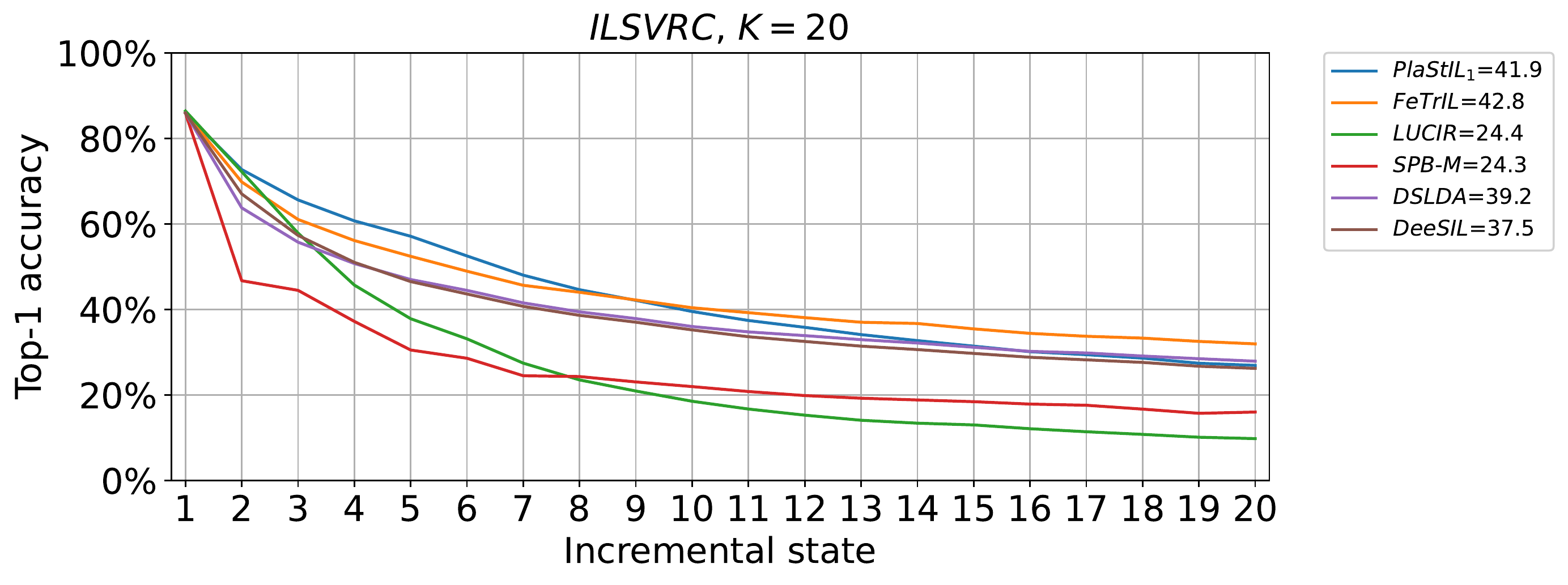}\\
 	\includegraphics[height=0.25\linewidth]{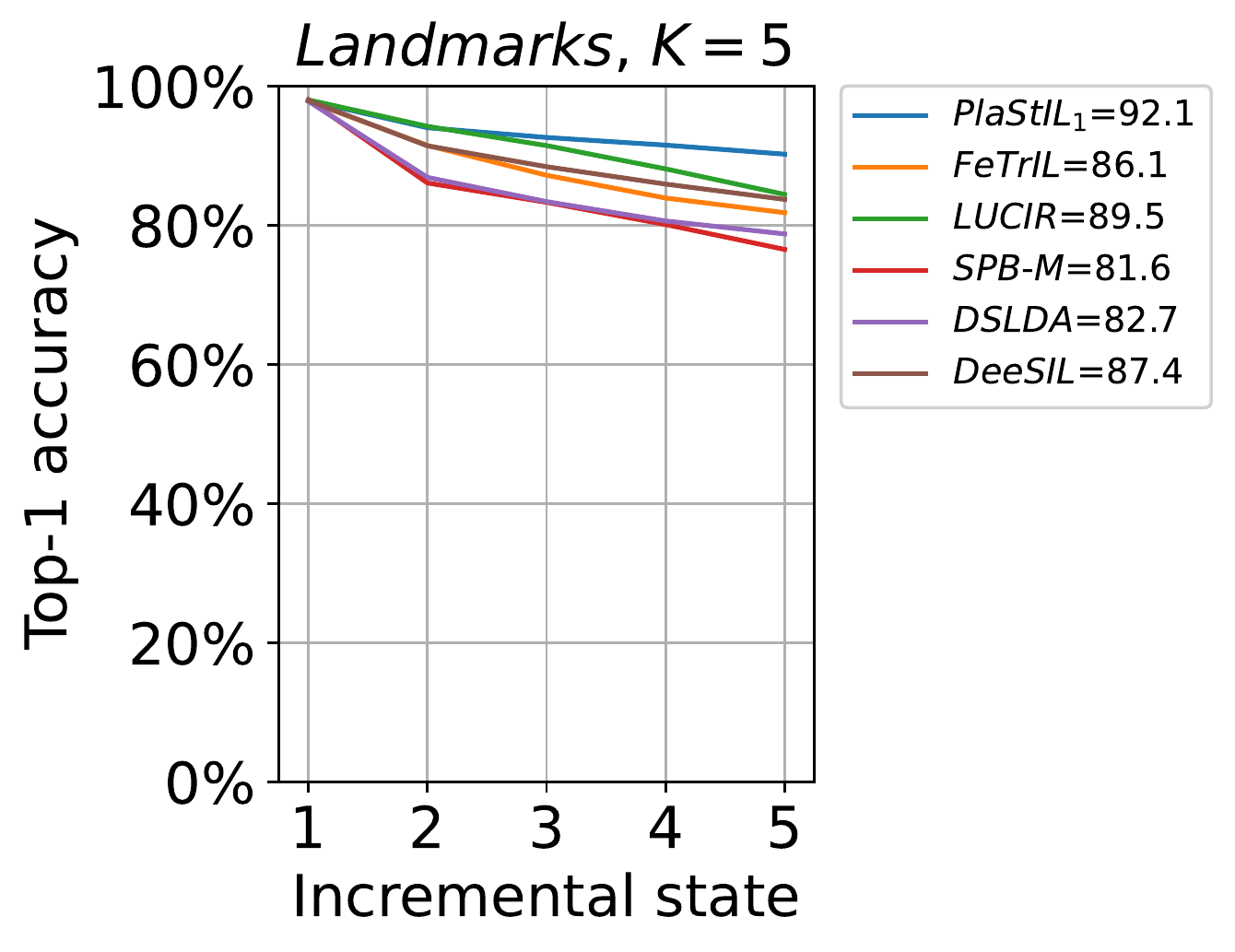}
	\includegraphics[height=0.25\linewidth]{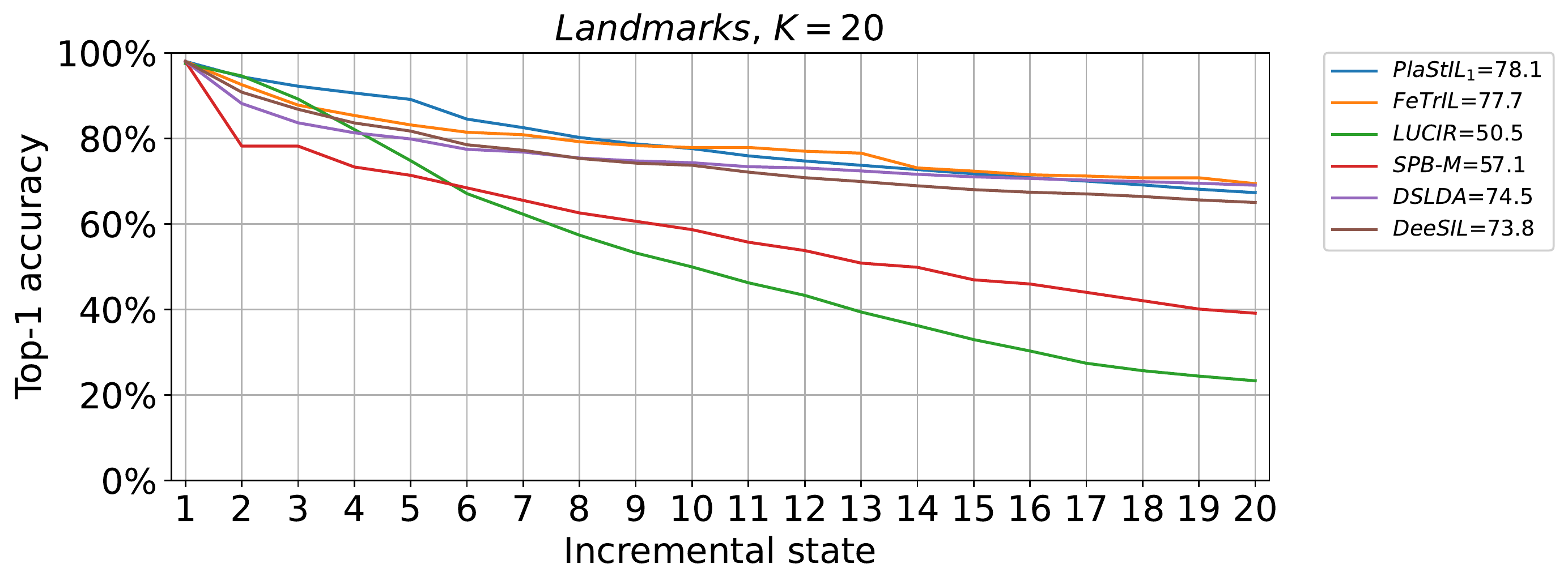}\\
 	\includegraphics[height=0.25\linewidth]{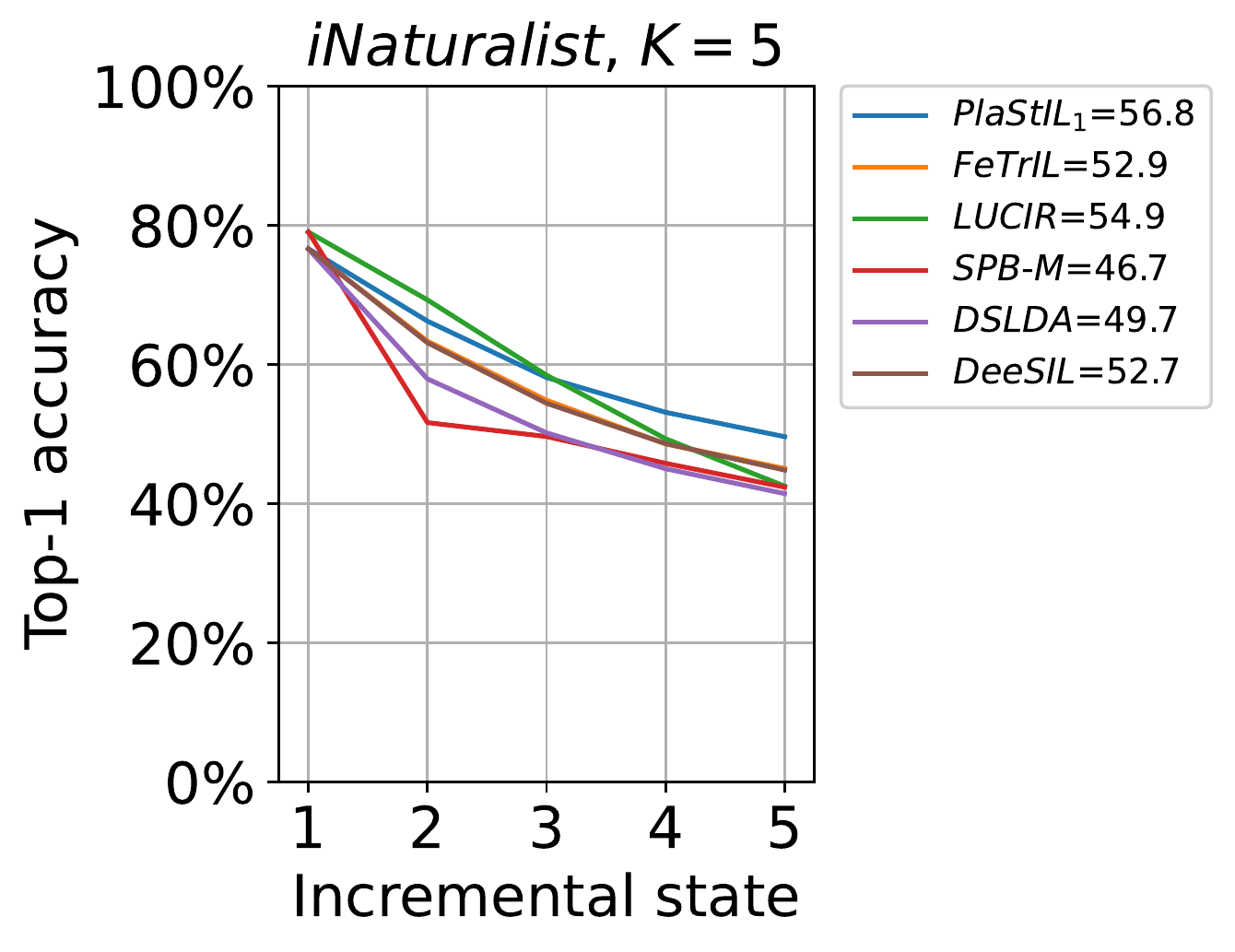}
	\includegraphics[height=0.25\linewidth]{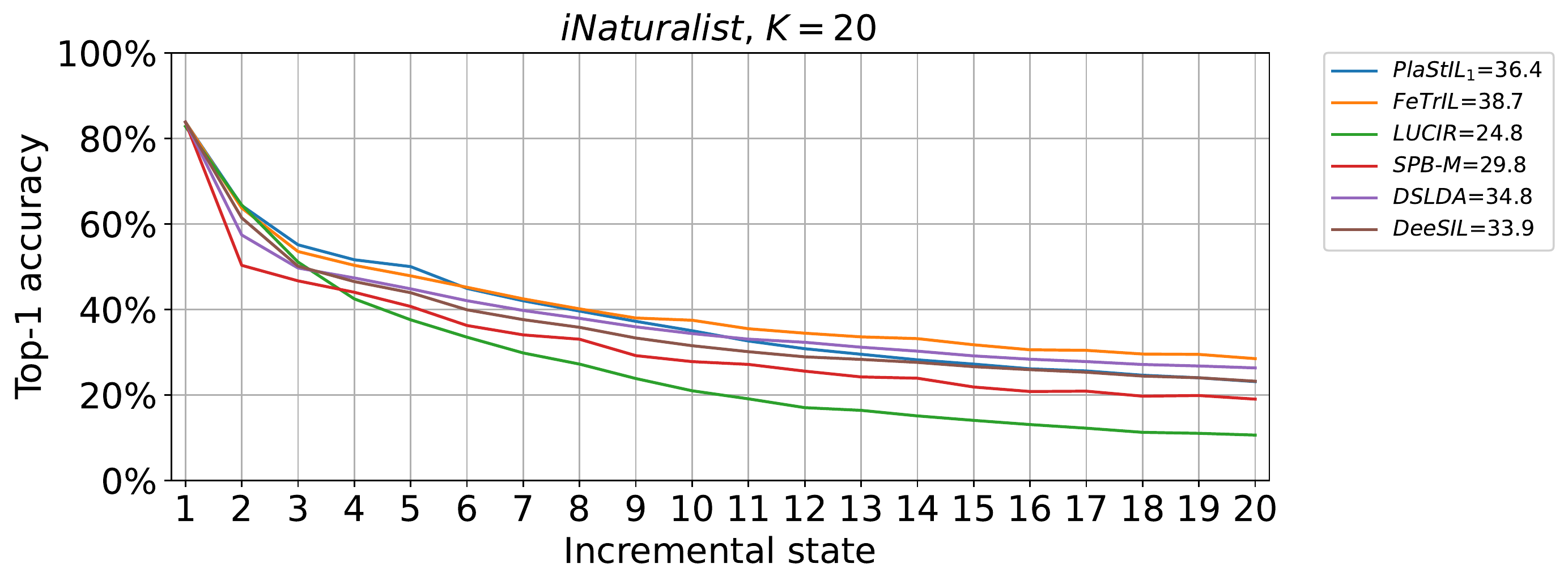}

	\caption{{Incremental accuracy across all states for} $K=5$ and $K=20$. {Plots for} $K=10$ {are presented in Figure}~\ref{fig:accuracy} {in the main paper. Plots are presented for the best methods from Table}~\ref{tab:main_results}. {\textit{Best viewed in color.}}}
	\label{fig:complete_acc}
\end{figure*}

\end{document}